\documentclass[aps,prf,reprint,amsmath,amssymb,onecolumn]{revtex4-2}
\usepackage{graphicx} 
\usepackage{tikz}
\usetikzlibrary{positioning, arrows.meta, shapes.geometric}
\usepackage{subcaption}
\usepackage{soul}
\usepackage{listings}
\usepackage{xcolor}
\usepackage{booktabs}
\usepackage{multirow}
\usepackage{array}
\usepackage[ruled,vlined]{algorithm2e}
\usepackage{xcolor}
\usepackage{float}    
\usepackage{makecell} 
\usepackage{footnote}
\usepackage{CJKutf8}
\usepackage{algorithm2e}  
\usepackage{dirtree}
\usepackage{forest}
\usepackage[margin=2.5cm]{geometry}

\RestyleAlgo{ruled}
\SetAlgoLined
\SetKwComment{Comment}{\color{blue}//}{}
\SetKwInput{KwInput}{\textbf{Input}}
\SetKwInput{KwOutput}{\textbf{Output}}
\SetKwInput{KwParams}{\textbf{Parameters}}
\SetKwFor{For}{\textbf{for}}{do}{\textbf{end}}
\SetKwIF{If}{ElseIf}{Else}{\textbf{if}}{then}{\textbf{else if}}{\textbf{else}}{\textbf{end}}
\SetKwFunction{Function}{Function}
\SetKwFunction{Try}{try}
\SetKwProg{Fn}{Function}{:}{end}
\usepackage[most]{tcolorbox}
\tcbset{
  colback=white,      
  colframe=black,     
  boxrule=0.5pt,      
  arc=2pt,            
  left=4pt, right=4pt, top=4pt, bottom=4pt, 
}

\begin{document}

\title{\texttt{Engineering.ai}: A Platform for Teams of AI Engineers in Computational Design}

\author{Ran Xu (\begin{CJK*}{UTF8}{gbsn}徐冉\end{CJK*})}
\thanks{These authors contributed equally to this work.}
\affiliation{
Faculty for Aerospace Engineering and Geodesy, University of Stuttgart, Stuttgart, Germany
}%

\author{Yupeng Qi (\begin{CJK*}{UTF8}{gbsn}亓宇鹏\end{CJK*})}
\thanks{These authors contributed equally to this work.}
\email{qiyupeng99@gmail.com}
\thanks{Corresponding author}
\affiliation{
Cluster of Excellence SimTech, University of Stuttgart, Stuttgart, Germany
}%

\author{Jingsen Feng (\begin{CJK*}{UTF8}{gbsn}冯晶森\end{CJK*})}
\affiliation{
Faculty of Environment, Science and Economy, University of Exeter, Exeter EX4 4QF, United Kingdom
}%

\author{Xu Chu (\begin{CJK*}{UTF8}{gbsn}初旭\end{CJK*})}
\affiliation{
Faculty of Environment, Science and Economy, University of Exeter, Exeter EX4 4QF, United Kingdom
}%
\affiliation{
University of Stuttgart, Stuttgart, Germany
}%

\begin{abstract}

In modern engineering practice, human engineers collaborate in specialized teams to design complex products, with each expert completing their respective tasks while communicating and exchanging results and data with one another. While this division of expertise is essential for managing multidisciplinary complexity, it demands substantial development time and cost. Recently, we introduced OpenFOAMGPT (1.0, 2.0) \cite{Pandey2025OpenFOAMGPT,Feng2025OpenFOAMGPT2}, which functions as an autonomous AI engineer for computational fluid dynamics, and \texttt{turbulence.ai}, which can conduct end-to-end research in fluid mechanics draft publications and PhD theses. Building upon these foundations, we present \texttt{Engineering.ai}, a platform for teams of AI engineers in computational design. The framework employs a hierarchical multi-agent architecture where a Chief Engineer coordinates specialized agents consisting of Aerodynamics, Structural, Acoustic, and Optimization Engineers, each powered by Large Language Model (LLM) with domain-specific knowledge. Agent-agent collaboration is achieved through file-mediated communication for data provenance and reproducibility, while a comprehensive memory system maintains project context, execution history, and retrieval-augmented domain knowledge to ensure reliable decision-making across the workflow. 
The system integrates FreeCAD, Gmsh, OpenFOAM, CalculiX, and Python-based Brooks-Pope-Marcolini (BPM) acoustic analysis, enabling parallel multidisciplinary simulations while maintaining computational accuracy. 
The framework is validated through UAV wing optimization, where agents autonomously evaluated four NACA airfoils across Reynolds numbers ranging from $10^5$ to $10^6$ using coupled OpenFOAM CFD simulations and BPM acoustic analysis. Experimental results demonstrate a reduction in setup time (from weeks to hours), and a complete automation of the CAD-CAE-optimization pipeline. This work demonstrates that agentic-AI-enabled AI engineers has the potential to perform complex engineering tasks autonomously, advancing the transition toward AI-assisted engineering design and analysis. Remarkably, the automated workflow achieved a 100\% success rate across over 400 parametric configurations, with zero mesh generation failures, solver convergence issues, or manual interventions required, validating that the framework is trustworthy. This work demonstrates that agentic-AI-enabled AI engineers have the potential to perform complex engineering tasks autonomously.
\end{abstract}

\maketitle

\section{Introduction}

The engineering design process has remained fundamentally unchanged for decades. Human engineers use computational tools as sophisticated calculators, manually setting up simulations, interpreting results, and making design decisions. While Computer-Aided Engineering (CAE) tools have become increasingly powerful, they still require extensive human expertise to operate effectively. High-fidelity simulations can demand weeks of iterative setup and debugging, and multidisciplinary analyses often operate in isolation because different software tools are difficult to integrate. This status quo not only limits productivity but also constrains the scope of problems that can be explored. 

In recent years, the emergence of large language models (LLMs) has elevated AI's generality and cognitive capabilities to significantly enhanced levels. These models can understand ambiguous natural language requirements and provide detailed responses with minimal human guidance~\cite{openai2023gpt4,anthropic2024claude3}. Recent comparative studies have demonstrated LLMs' proficiency in scientific computing tasks. Jiang et al.~\cite{jiang2025deepseek} evaluated DeepSeek, ChatGPT, and Claude on PDE-based problems and scientific machine learning, revealing that reasoning-optimized models consistently outperform non-reasoning counterparts in solving challenging computational problems. Building upon these capabilities, researchers have begun exploring how to apply LLMs in engineering domains. The key paradigm to realize this leap is an LLM-driven multi-agent system, in which multiple specialized AI agents work together to tackle complex tasks~\cite{wang2023survey,xi2023rise}.

The development of multi-agent frameworks has demonstrated remarkable potential for automating complex scientific workflows. Recent frameworks such as MetaGPT~\cite{Hong2024MetaGPT} have encapsulated human team collaboration patterns into prompt strategies, assigning roles to agents (planner, coder, tester, etc.) that communicate with each other and verify outputs. This structured "role-playing" significantly improves the coherence and reliability of solving complex multi-step problems. Multi-agent "AI Scientist" platforms have taken this concept further by autonomously generating hypotheses, designing and executing experiments, and even writing scientific papers without human assistance. For example, Google's AI Co-Scientist~\cite{Gottweis2025AICoScientist} uses a multi-agent debate and optimization strategy to propose novel ideas in biomedical research, some of which were later validated experimentally. Yamada et al.'s AI Scientist-v2~\cite{Yamada2025AIScientistV2} achieves workshop-level automated scientific discovery through agentic tree search, capable of handling more complex experimental designs. In computational physics, Yang et al.\cite{yang2025large} demonstrated LLM-driven turbulence model development, treating the model as an equal research partner to synthesize physically interpretable approaches through iterative reasoning. Similarly, the Sakana AI Scientist~\cite{lu2024aiscientist} project achieved a fully automated research pipeline from hypothesis generation to paper writing, and produced what is claimed to be the first AI-generated research paper to pass peer review. Aygün et al.~\cite{aygun2025ai} developed an AI system combining LLMs with tree search to write expert-level empirical software across diverse scientific domains including bioinformatics, epidemiology, and geospatial analysis, demonstrating that AI can systematically explore solution spaces to discover novel methods outperforming human-developed approaches.

Multi-agent AI systems have achieved remarkable progress in computational fluid dynamics (CFD). The high complexity of traditional CFD workflows has long constrained their widespread adoption, which precisely provides broad opportunities for AI-driven automation. Our early OpenFOAMGPT 1.0 ~\cite{Pandey2025OpenFOAMGPT, wang-2025} demonstrated that with proper prompting, an LLM can configure and run an OpenFOAM-based CFD case using only natural language instructions. It could build a typical flow case from scratch, adjust boundary conditions or turbulence models on demand, even translate setups between different solvers, and it employed an iterative loop to catch and correct errors. Dong et al.~\cite{dong2025finetuning} explored domain-specific fine-tuning by training Qwen2.5-7B on NL2FOAM, a dataset of 28,716 natural language-to-OpenFOAM configuration pairs, achieving 88.7\% solution accuracy and 82.6\% first-attempt success rate while outperforming larger general-purpose models. To systematically evaluate LLM performance in CFD tasks, Somasekharan et al.~\cite{somasekharan2025cfd} developed CFDLLMBench, a benchmark suite testing models on graduate-level knowledge, numerical reasoning, and workflow implementation capabilities.

As our requirements grew, we introduced a multi-agent architecture to improve robustness and extend to more complex scenarios. OpenFOAMGPT~2.0~\cite{Feng2025OpenFOAMGPT2} introduced four specialized agents consisting of Pre-processing, Prompt Generation, Simulation, and Post-processing, forming an end-to-end autonomous CFD assistant, and demonstrated that an LLM-driven agent system can meet the high-precision standards required in scientific computing workflows. This approach has inspired parallel developments. MetaOpenFOAM~\cite{Chen2024MetaOpenFOAM} combined multiple GPT-based agents to break down the CFD workflow into subtasks such as mesh generation, solver execution, and post-processing; CFDagent~\cite{Yao2025CFDagent} integrated three GPT-4-driven agents responsible for geometry generation, meshing, solver execution, and result analysis, exploring LLM-agent collaboration that combines geometric processing and visualization. Yue et al.~\cite{yue2025foamagent} developed Foam-Agent, a hierarchical multi-agent framework that achieved 83.6\% success rate in automating CFD simulations through dependency-aware file generation and iterative error correction mechanisms, which was later enhanced in Foam-Agent 2.0~\cite{yue2025foamagent20} with a composable service architecture using Model Context Protocol, achieving 88.2\% success rate on 110 simulation tasks. Fan et al.~\cite{fan2025chatcfd} introduced ChatCFD, an LLM-driven agent utilizing DeepSeek models for end-to-end OpenFOAM automation, achieving 82.1\% operational success rate across 205 benchmark cases through domain-specific structured reasoning. Comprehensive evaluations by Wang et al.~\cite{wang2025evaluations} across conventional CFD problems, physics-informed neural networks, and ill-conditioned systems reveal that while reasoning LLMs demonstrate superior performance in leveraging existing knowledge, their autonomous knowledge creation capabilities remain an area for improvement.

Beyond fluid dynamics, LLM-driven automation has extended across diverse engineering disciplines, demonstrating the broad applicability of this paradigm. Park et al.'s generative agents~\cite{park2023generative} demonstrated that AI agents can simulate believable human behavior, providing theoretical foundations for multi-agent collaboration applicable across domains. In automotive design, Elrefaie et al. proposed a multi-agent design assistant~\cite{Elrefaie2025DesignAgents}, which uses an LLM together with vision models to interpret design sketches and generate 3D car models.

The CAD domain has witnessed transformative advances through LLM integration. Khan et al.~\cite{khan2024text2cad} developed Text2CAD for generating sequential parametric CAD models from natural language prompts, Li et al.~\cite{li2025llm4cad} introduced LLM4CAD enabling 3D design generation from combined text-image inputs, and Xu et al.~\cite{xu2024cadmllm} presented CAD-MLLM integrating diverse input modalities including point clouds for comprehensive CAD automation. M\"oltner et al.~\cite{moltner-2025} developed a comprehensive framework for automated simulation model generation and validation in mechanical engineering, demonstrating that LLMs can effectively identify incorrect multibody dynamics models with high accuracy. Zhang et al.~\cite{zhang2025llmpso} explored LLMs for parametric shape optimization, developing the LLM-PSO framework that leverages in-context learning to optimize engineering designs across multiple flow-related problems, including airfoil lift-to-drag maximization and heat exchanger thermal resistance minimization. 

In the domain of AI Scientist, we have developed turbulence.ai~\cite{Feng2025turbulenceai}, a system capable of autonomously conducting the entire research process—from problem formulation to simulation and analysis. It can design its own simulation plans based on high-level research questions, retrieve and adapt relevant knowledge to inform setup, execute simulations while dynamically handling any required errors or modifications, and ultimately interpret the results to generate scientific insights.

Despite these substantial advances, significant challenges remain for AI agents to fully assume the role of autonomous engineers. In complex scenarios, iterative self-correction and trustworthiness cannot be guaranteed, with systems lacking transparency and explainability in their decision-making processes. Current AI workflows struggle to coordinate multiple heterogeneous software tools effectively, failing to achieve true toolchain integration. Moreover, existing systems are unable to autonomously construct and solve multidisciplinary coupled analyses, lacking the capability to understand interactions between different physical domains. In structural mechanics specifically, while AutoFEA~\cite{hou2025autofea} has demonstrated the viability of LLM-assisted finite element analysis workflows by combining graph neural networks with language models for automated code generation, comprehensive automation from natural language to optimized structural designs remains an open challenge. The insufficient cross-domain knowledge transfer capability limits AI's ability to learn from experiences in one engineering field and apply them to others. Finally, error recovery mechanisms remain inadequate, as systems often cannot autonomously recover or find alternative solutions when encountering unforeseen failure modes.

This paper introduces \texttt{Engineering.ai}, a platform for teams of AI engineers that addresses these challenges through comprehensive integration across multiple engineering domains. The system has the potential to independently conduct complex engineering tasks with minimal human intervention. \texttt{Engineering.ai} integrates multiple LLM-driven agents into the traditional CAE workflow, from parametric CAD modeling and mesh generation to multidisciplinary simulation, autonomous optimization, and result interpretation, implementing these capabilities through a closed-loop, self-correcting architecture. We demonstrate near-autonomous engineer-level operational and decision-making capabilities, leveraging open-source tools (FreeCAD for geometry~\cite{freecad2024}, Gmsh for meshing~\cite{geuzaine2009gmsh}, OpenFOAM for fluid dynamics~\cite{weller1998tensorial,chu2019direct}, CalculiX for structural analysis) under the guidance of an LLM controller. The system can interpret natural language requirements, decompose them into technical specifications, select appropriate computational methods, detect and correct errors, autonomously explore design spaces, and discover optimal solutions balancing multiple competing objectives. Through multi-agent collaborative decision-making, iterative error correction, and contextual knowledge retrieval mechanisms, \texttt{Engineering.ai} ensures scientific rigor, and within the experimental scope of this study, the system was able to produce highly accurate and reproducible results.

\section{Methodology: Leader-Orchestrated Multi-Agent Engineering Framework}

The Engineering.ai framework is powered by Gemini 2.5 pro. Traditional engineering design follows a well-established collaborative paradigm. A typical engineering project is organized hierarchically, where a project manager or lead engineer oversees the entire workflow while specialized engineers focus on distinct aspects of the design. Some analyze physical behaviors, others evaluate structural integrity, and additional team members optimize performance metrics or assess environmental impacts. These specialists work in coordinated cycles of design, analysis, and refinement. Initial concepts are developed, then passed to analysts who evaluate feasibility through computational simulations or physical testing. Results flow back to designers who refine the geometry, triggering subsequent analysis iterations. Throughout this process, engineers communicate through meetings, technical reports, and data files exchanged between different software tools. A senior engineer or design lead synthesizes inputs from various disciplines, resolves conflicting requirements, and makes final decisions balancing multiple competing objectives. While this human-centered approach has proven effective for decades, it suffers from inherent inefficiencies. Sequential dependencies create bottlenecks, manual data transfer between tools introduces errors and delays, significantly increasing development time and cost.

\subsection{AI Engineering Team Architecture}

\texttt{Engineering.ai} introduces a hierarchical engineering team structure inspired by successful human engineering organizations. The core innovation is leveraging our engineering know-how to seamlessly integrate LLMs with engineering software. By deeply understanding both engineering practices and LLM capabilities, we design autonomous collaboration protocols where specialized agents communicate through structured interfaces, coordinate multidisciplinary analyses, and collectively solve complex design problems as human engineering teams do.

\begin{figure}[htbp]
    \centering
    \includegraphics[width=1\linewidth]{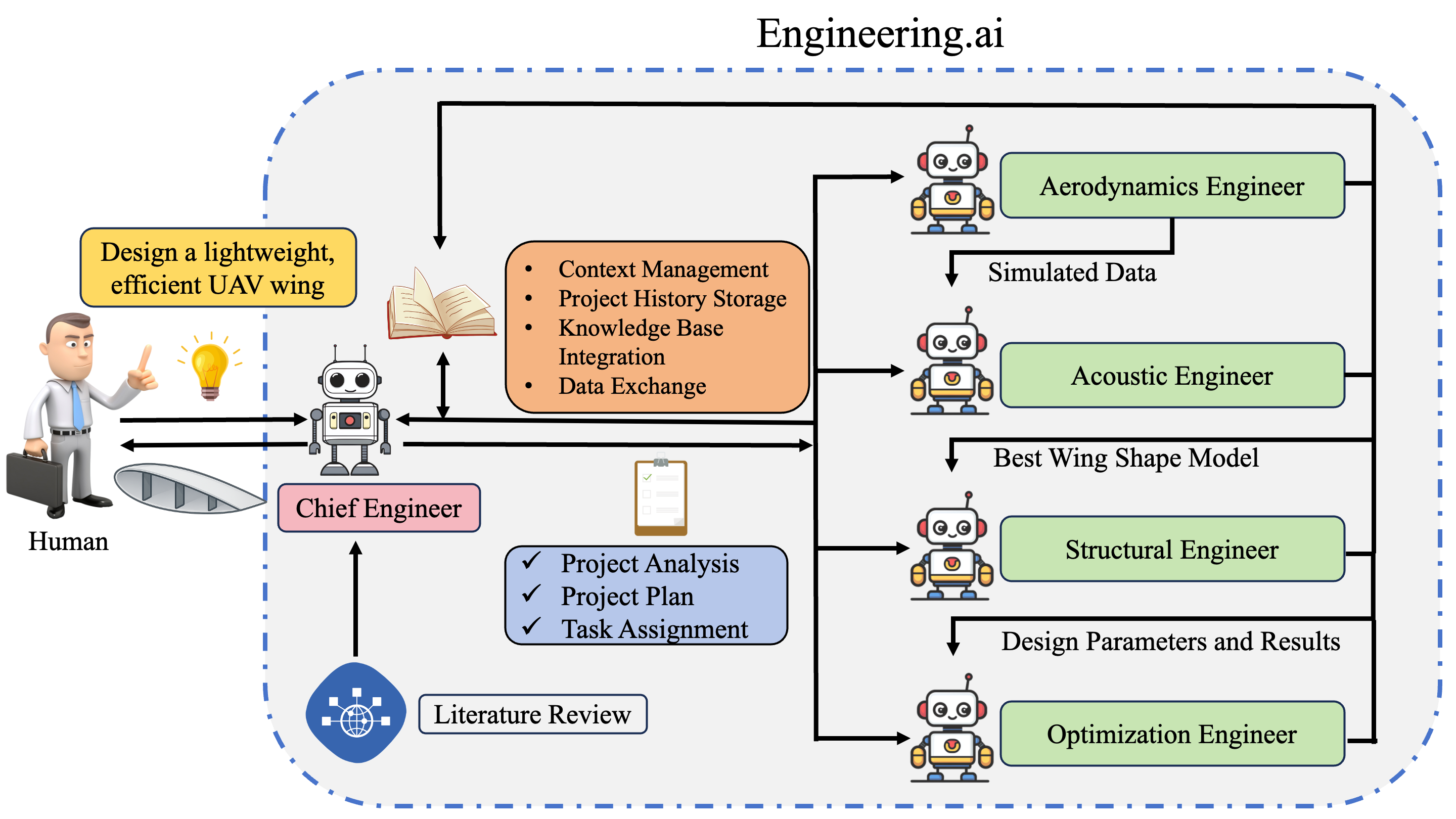}
    \caption{Engineering.ai}
    \label{fig:Engineering.ai}
\end{figure}

As shown in Figure \ref{fig:Engineering.ai}, the architecture presents a complete engineering design ecosystem where human users interact with the system through natural language, the Chief Engineer serves as the central coordinator managing the entire team, specialized agents perform their respective duties, and the memory system provides continuous knowledge support. The left side of Figure \ref{fig:Engineering.ai} demonstrates how human users can initiate the entire design process through simple natural language descriptions (e.g., "Design a lightweight and efficient UAV wing"). The Chief Engineer at the center is not merely a simple task dispatcher but an intelligent coordinator with engineering judgment capabilities, able to understand overall project objectives, identify potential technical conflicts, optimize resource allocation, and ensure effective integration of work outputs from various agents. The four specialized engineering agents on the right, Aerodynamics Engineer, Acoustic Engineer, Structural Engineer, and Optimization Engineer, each possess deep domain knowledge and access to professional tools. They operate with professional autonomy while maintaining close collaboration with colleagues, mirroring the dynamics of human engineering teams.

It is important to note that the engineering team concept we introduce transcends the specific disciplines demonstrated in this work. While we instantiate the framework with Aerodynamics, Structural, Acoustic, and Optimization Engineers, the hierarchical multi-agent design is not limited to these disciplines. The system's extensibility lies in the coordination mechanisms and standardized communication protocols among agents. For instance, thermal engineers could be added through interfaces to heat transfer solvers, manufacturing engineers through CAM software integration, or control system engineers via simulation platforms like Simulink. Each new agent inherits the framework's core capabilities including autonomous error recovery, structured communication protocols, and Chief Engineer coordination, requiring only domain-specific tool integration and knowledge base customization. This concept allows Engineering.ai to address diverse multidisciplinary problems across different industries by adding appropriate specialist agents.

\subsection{Chief Engineer: Project Management and Coordination}

The Chief Engineer functions as the intelligent workflow orchestrator, employing advanced LLM capabilities to understand customer requirements, plan projects, and coordinate the engineering team. Upon receiving customer specifications, the Chief Engineer conducts a requirements analysis, extracting key objectives, constraints, and success criteria. The agent then decomposes the high-level task into a structured workflow, identifying task dependencies and opportunities for parallel execution. For example, in the actual UAV wing optimization project, after querying the knowledge base for relevant airfoil research, the Chief Engineer creates a project plan that includes multiple NACA airfoils, different flow velocities, and various angles of attack. Each design proposal is evaluated against physical constraints, material properties, and computational resource limits, ensuring that the proposed designs remain within the bounds of physical feasibility and computational tractability.

he Chief Engineer's core functionalities center on task planning and delegation. The agent decomposes user requirements into specific engineering tasks (e.g., ``Aerodynamics Engineer: analyze NACA 4412 at $\mathrm{Re}=3.5\times10^{5}$'', ``Structural Engineer: perform FEA with aerodynamic loads from previous step''), assigns them to appropriate specialized engineers in dependency order, and monitors workflow execution through structured logging. Beyond workflow orchestration, the Chief Engineer performs critical technical decisions: determining simulation parameter matrices (combinations of airfoils, velocities, and angles of attack), managing data flow between agents by specifying file paths for inter-agent data exchange (e.g., instructing the Structural Engineer to read aerodynamic loads from \texttt{cfd\_results/forces.dat}), and resolving inter-disciplinary coupling issues when multiple physics domains interact. This intelligent coordination ensures systematic design-space exploration and seamless integration of multi-disciplinary simulation results.

\subsection{Specialized Engineering Agents}

The four specialized engineering agents shown on the right side of Figure \ref{fig:Engineering.ai} represent the core technical capabilities of \texttt{Engineering.ai}. Unlike traditional software interfaces, these agents integrate domain-specific expertise with autonomous decision-making and collaborative abilities, enabling them to understand physical principles, make engineering judgments, and learn from errors.

\textbf{Aerodynamics Engineer}: The agent integrates OpenFOAMGPT 2.0~\cite{Feng2025OpenFOAMGPT2} capabilities to bridge natural language with technical engineering requirements. It automatically configures turbulence model parameters~\cite{spalart1992one,menter1994two}, boundary conditions, and multiple post-processing functions while accounting for the complexities of turbulence simulations. The agent automatically generates meshes via Gmsh~\cite{geuzaine2009gmsh} and validates mesh quality using OpenFOAM's~\cite{weller1998tensorial,jasak1996error} built-in checking and optimization tools. The agent implements security-aware configuration generation by precomputing all mathematical expressions, thereby avoiding runtime computations that could pose security risks in containerized environments. Furthermore, it converts raw simulation data into publication-quality figures.

\textbf{Acoustic Engineer}: This agent employs the Brooks-Pope-Marcolini (BPM) model~\cite{Brooks1989, Lau2017}, automatically extracting key parameters such as boundary layer thickness, displacement thickness, and momentum thickness from CFD results, and calculates contributions from five noise mechanisms: turbulent boundary layer trailing edge noise (TBL-TE), laminar boundary layer vortex shedding noise (LBL-VS), separation flow noise, blunt trailing edge vortex shedding noise, and tip vortex noise. The agent performs comprehensive spectral analysis within the audible range (20 Hz to 20 kHz), calculating Sound Pressure Level (SPL), Overall Sound Pressure Level (OASPL), and frequency-weighted metrics (dBA, dBC). It generates directivity patterns that illustrate noise radiation characteristics across multiple observation angles.

\textbf{Structural Engineer}: This agent manages the complete CAD-to-FEA workflow leveraging open-source tools FreeCAD~\cite{freecad2024}, Gmsh~\cite{geuzaine2009gmsh}, and CalculiX~\cite{dhondt2004calculix}. Starting from natural language geometry descriptions or STEP files, the agent uses FreeCAD to generate parametric 3D models with precise geometric features (airfoil profiles, internal ribs and spars, shell structures). The geometry is then meshed using Gmsh, which generates adaptive tetrahedral elements~\cite{shewchuk1996triangle,field1988laplacian} with local refinement near stress concentrations and geometric discontinuities. For finite element analysis, the agent interfaces with CalculiX, configuring material properties (elastic modulus, Poisson's ratio, density), applying boundary conditions (fixed supports, symmetry planes), and loading conditions derived from aerodynamic simulations. The agent automatically extracts stress distributions, displacement fields, and structural mass from CalculiX output files (.frd, .dat), enabling autonomous structural evaluation and design iteration~\cite{hughes1987finite,zienkiewicz2000finite}. The agent supports batch processing capabilities, autonomously executing multiple design configurations in parallel to enable comprehensive parameter sweeps and design space exploration.

\textbf{Optimization Engineer}: Serving as the system's data intelligence hub, this agent autonomously analyzes user requirements and problem characteristics to select appropriate optimization strategies. Based on the task at hand (e.g., "find lightweight designs", "explore design space efficiently", "balance multiple objectives"), the agent intelligently chooses from a portfolio of methods including surrogate modeling, multi-objective optimization, sensitivity analysis, and uncertainty quantification. For example, when presented with a structural optimization task involving large-scale parameter sweeps, the agent may autonomously select Gaussian Process regression for surrogate modeling combined with Bayesian optimization for design space exploration, identifying optimal configurations that balance structural performance and weight constraints.

\subsection{Data Management and Knowledge Integration}

The framework implements a structured data management system that enables knowledge retention, information exchange, and continuous knowledge support:

\textbf{Context Management}: The Chief Engineer maintains the current project state through structured LLM prompts, including active design parameters, simulation results, and task dependencies. When delegating tasks to specialized engineers, the Chief Engineer selectively provides relevant context (e.g., aerodynamic loads for structural analysis, geometric constraints for mesh generation) while filtering out irrelevant details to maintain clarity and focus.

\textbf{Project History Storage}: The system stores complete project execution records in file-based formats (JSON, CSV, VTK), capturing design decisions, simulation results, and optimization trajectories. Each workflow stage generates structured output files that serve as both archival records and inputs for subsequent stages. This enables result reproducibility, facilitates design iteration, and supports post-analysis of design evolution.

\textbf{Knowledge Base Integration}: Domain knowledge including material properties, design rules, physics principles, and best practices is accessible through retrieval-augmented generation (RAG) mechanisms. Agents query this knowledge base when making design decisions or configuring simulations, enabling evidence-based engineering choices grounded in established literature and expert knowledge.

\textbf{Data Exchange}: Specialized engineers communicate through file-based data exchange. For example, the Aerodynamics Engineer generates aerodynamic load distributions (forces.dat, pressure\_field.json), which the Structural Engineer reads to configure FEA boundary conditions. The Optimization Engineer aggregates results from all engineers (stress\_analysis.json, weight\_data.json, acoustic\_spectrum.dat) to perform multi-objective optimization. This file-mediated architecture ensures data provenance, enables parallel execution, and facilitates debugging and result verification.

\begin{tcolorbox}[title={Project Storage}]
\begin{small}
  \dirtree{%
  .1 project/.
  .2 pipeline.log.
  .2 airfoil/.
  .3 idea.json.
  .3 result.md.
  .3 aerodynamics\_plan.md.
  .3 acoustics\_plan.md.
  .3 multi\_case\_analysis/.
  .4 aerodynamic\_data.csv, plot\_aerodynamic\_analysis.png.
  .4 acoustic\_data.csv, plot\_acoustic\_analysis.png.
  .2 sim\_NACA0012\_25ms\_aoa0/ \textit{(CFD case)}.
  .3 mesh.md, airfoil.geo, airfoil.msh.
  .3 constant/, system/, 0/, Allrun.
  .3 50/, 100/, 150/, 200/, 250/, 300/.
  .3 acoustics\_data/.
  .4 flow\_field.json, bpm\_input.json, boundary\_layer.json.
  .3 postProcessing/.
  .4 forceCoeffs/0/coefficient.dat.
  .4 surfaces/\{50,100,...,300\}/.
  .5 p\_airfoilSurface.raw, U\_airfoilSurface.raw.
  .4 integrated/ \textit{(Comprehensive Results)}.
  .5 force\_coefficients.csv.
  .5 boundary\_layer.csv.
  .5 cp\_data.csv.
  .5 figures/.
  .6 openfoam\_3000\_p\_field.png, openfoam\_3000\_U\_field.png.
  .5 acoustics/.
  .6 acoustic\_metrics.csv \textit{--- OASPL, SPL, dBA}.
  .6 third\_octave\_spectrum.csv.
  .3 VTK/.
  .4 openfoam.vtm.series.
  .4 openfoam\_3000.vtm.
  .4 openfoam\_3000/.
  .2 \textit{+ 11 more simulation cases}.
  }
\end{small}
\end{tcolorbox}

\subsection{Agent-Agent Collaboration and Communication}

The framework implements sophisticated collaboration mechanisms that go beyond simple message passing, enabling true collective intelligence:

\textbf{Collaborative Problem Solving}: For multi-physics problems, agents form temporary coalitions coordinated by the Leader. In aeroacoustic optimization, the Fluid Dynamics and Acoustic Agents work in tight coupling—the former predicting flow fields while the latter computes noise propagation. They exchange intermediate results through file-based data exchange, enabling rapid iteration with Chief Engineer coordination.

\textbf{Knowledge Synthesis}: Agents actively share domain insights through the knowledge base. When the FEA Agent discovers that certain rib configurations consistently fail under fatigue loading, this finding is documented in the project history and can be referenced by the Chief Engineer when planning future designs. The Optimization Agent can access these documented failures to avoid exploration of known failure modes.

\textbf{Parallel Simulation Scheduling}: Advanced scheduling and resource management algorithms are implemented to maximize computational throughput while maintaining system stability. Independent simulations are distributed across available cores through task-level parallelism, data-level parallelism is achieved within a single OpenFOAM simulation using MPI, and pipeline parallelism overlaps I/O operations with computation. The optimized hybrid scheduling enables intelligent task partitioning by analyzing simulation parameters to identify parallelization opportunities. A dynamic priority queue is maintained to adjust task order based on estimated completion time and resource availability. For UAV wing design applications, maximum parallelism is achieved through simulation-stage separation, allowing up to four concurrent Docker calls.

\textbf{Error Recovery and Checkpointing}: Fault-tolerance capabilities are provided for long-running computational workflows. At stage boundaries, the agent automatically creates checkpoints by serializing the complete process to disk in JSON format. These checkpoints enable workflow recovery from the last successfully completed stage, which is critical for computations that may run for days or weeks. The recovery mechanism includes immediate retries with exponential backoff for transient failures (e.g., network timeouts, API rate limits), and garbage collection with memory cleanup before retrying in cases of resource exhaustion. The system maintains detailed fault logs to inform both immediate recovery decisions and long-term system improvements.

\begin{algorithm}[t]
\caption{Intelligent Error Recovery with Domain-Specific Strategies}
\label{algorithm_recovery}
\KwInput{Pipeline State $\mathcal{S}$, Error Event $\epsilon$}
\KwOutput{Recovered State $\mathcal{S}'$, Success Flag}
\KwParams{$max\_retries = 3$, $checkpoint\_interval = 10$ stages}

\tcp{Checkpoint Management}
\If{$\mathcal{S}$.stage mod $checkpoint\_interval == 0$}{
    Serialize state: $\mathcal{S} \rightarrow$ gzip(pickle($\mathcal{S}$))\;
    Save metadata: \{phase, progress, MD5 hash, timestamp\}\;
    Maintain sliding window of 10 checkpoints\;
}

\tcp{Error Diagnosis via Log Analysis}
$error\_class$ $\leftarrow$ ParseDockerLogs($\epsilon$.logs)\;
\Switch{$error\_class$}{
    \Case{MeshConversionFailure}{
        $strategy$ $\leftarrow$ \{action: regenerate\_mesh,
                     params: refinement $\times$ 0.8\}\;
    }
    \Case{SolverDivergence}{
        $strategy$ $\leftarrow$ \{action: adjust\_relaxation,
                     pressure: 0.3, velocity: 0.2,
                     timeStep: current $\times$ 0.5\}\;
    }
    \Case{BoundaryConditionError}{
        $strategy$ $\leftarrow$ \{action: correct\_patches,
                     mapping: \{walls $\rightarrow$ wall,
                               front/back $\rightarrow$ empty\}\}\;
    }
    \Other{
        $strategy$ $\leftarrow$ DefaultRecovery()\;
    }
}

\tcp{Adaptive Retry with Exponential Backoff}
\For{$attempt \leftarrow 1$ \KwTo $max\_retries$}{
    $\mathcal{S}'$ $\leftarrow$ ApplyStrategy($\mathcal{S}$, $strategy$)\;
    \If{ExecuteSimulation($\mathcal{S}'$) succeeds}{
        ValidateIntegrity($\mathcal{S}'$)\;
        \Return{$\mathcal{S}'$, true}\;
    }
    \If{$attempt < max\_retries$}{
        $\mathcal{S}$ $\leftarrow$ LoadCheckpoint(last\_valid)\;
        AdjustStrategy($strategy$, $attempt$)\;
        Wait($2^{attempt}$ seconds)\;
    }
}
\Return{$\mathcal{S}$, false}\;
\end{algorithm}

Algorithm~\ref{algorithm_recovery} presents the checkpoint-based error recovery system implemented in \texttt{Engineering.ai}. The system automatically creates compressed checkpoints using gzip and pickle serialization at phase boundaries, maintaining up to 10 checkpoints with metadata including phase status, progress percentage, and MD5 hash for integrity verification. When Docker execution encounters errors such as mesh conversion failures, solver divergence, or boundary condition misconfigurations, the system diagnoses the specific error type through log parsing and applies targeted recovery strategies. For mesh-related failures, it reduces refinement parameters by 20\%; for solver divergence, it adjusts relaxation factors from 0.7 to 0.3 for pressure and 0.5 to 0.2 for velocity; for boundary errors, it automatically corrects patch types (walls to wall, front/back to empty). The retry mechanism allows up to 3 attempts, with automatic rollback to the last valid checkpoint if recovery fails, ensuring that the pipeline can resume from a known good state without losing progress.

\subsection{LLM-Driven Autonomous Workflow Management}

The framework's autonomy is achieved through sophisticated LLM-driven workflow orchestration that seamlessly integrates ideation, simulation, and analysis phases without human intervention. The \texttt{Engineering.ai} pipeline transforms natural language requirements into executable computational workflows through intelligent task decomposition and parallel execution strategies. This autonomous capability is realized through a hierarchical control architecture where the Chief Engineer coordinates specialized agents, each responsible for distinct computational domains yet operating within a unified execution framework.

Algorithm~\ref{algorithm_full} presents the complete autonomous engineering workflow as implemented in the \texttt{Engineering.ai} system. The algorithm orchestrates the entire process from initial requirements through parallel simulations to comprehensive analysis, demonstrating how LLM-driven intelligence enables truly autonomous computational engineering.

\begin{algorithm}[H]
\caption{LLM-Driven Multi-Agent Engineering Pipeline}
\label{algorithm_full}
\KwInput{Natural Language Requirements $R$}
\KwOutput{Optimized Design $\mathcal{D}_{opt}$, Performance Metrics $\mathcal{M}$}
\KwParams{$max\_parallel = 4$, solver = \texttt{simpleFoam}}

\tcp{PHASE 1: Literature Review and Experimental Design}
ChiefEngineer.AnalyzeLiterature($R$) $\rightarrow$ knowledge\_base\;
$matrix$ $\leftarrow$ LLM.GenerateExperiments(knowledge\_base)\;

\tcp{PHASE 2: Aerodynamic Analysis (OpenFOAM)}
\tcp{Parallel Simulation Setup}
\ForEach{$config \in matrix$ \textbf{in parallel}}{
    $mesh$ $\leftarrow$ Gmsh.AdaptiveMesh($config$.geometry)\;
    $BC$ $\leftarrow$ $(U\cos\alpha, U\sin\alpha, 0)$ at inlet\;
    $case$ $\leftarrow$ OpenFOAMCase($mesh$, $BC$, SpalartAllmaras)\;
}

\tcp{Data Analysis}
\ForEach{completed simulation}{
    \textbf{Aerodynamics:} $C_L$, $C_D$, $L/D$ from forces\;
}

\tcp{PHASE 3: Acoustic Analysis (BPM Model)}
\ForEach{completed simulation}{
    \textbf{Acoustics:} Extract $\delta^*$, $\theta$, $H$ for BPM\;
    \quad Compute SPL spectrum [100Hz, 10kHz]\;
    \quad Calculate OASPL and directivity\;
}

\tcp{PHASE 4: Multi-Objective Airfoil Selection}
$J_{aero} = 0.6 \cdot \frac{L/D}{(L/D)_{max}} + 0.4 \cdot (1 - \frac{OASPL - OASPL_{min}}{OASPL_{max} - OASPL_{min}})$\;
$\mathcal{D}_{selected}$ $\leftarrow$ $\arg\max(J_{aero})$\;
$\mathcal{F}_{aero}$ $\leftarrow$ Extract aerodynamic loads (Lift, Drag) for $\mathcal{D}_{selected}$\;

\tcp{PHASE 5: Structural Analysis (FreeCAD + Gmsh + CalculiX)}
$\mathit{struct\_params}$ $\leftarrow$ \{$\mathit{SPAR\_WIDTH}$, $\mathit{RIB\_THICKNESS}$, ...\}\;
$\mathit{struct\_matrix}$ $\leftarrow$ $\mathit{GenerateParameterSweep}$($\mathit{struct\_params}$)\; \tcp{432 configs}

\ForEach{$\mathit{config} \in \mathit{struct\_matrix}$ \textbf{in parallel}}{
    $\mathit{geometry}$ $\leftarrow$ $\mathit{FreeCAD.GenerateParametricWing}$($\mathcal{D}_{selected}$, $\mathit{config}$)\;
    $\mathit{mesh\_FEA}$ $\leftarrow$ $\mathit{Gmsh.TetMesh}$($\mathit{geometry}$) with local refinement\;

    $\mathit{BC\_fixed}$ $\leftarrow$ Root constraint\;
    $\mathit{BC\_loads}$ $\leftarrow$ Apply $\mathcal{F}_{aero}$ to wing surface \tcp{Inherit CFD loads}\;

    $\mathit{FEA\_case}$ $\leftarrow$ $\mathit{CalculiX.Setup}$($\mathit{mesh\_FEA}$, $\mathit{BC\_fixed}$, $\mathit{BC\_loads}$, Al7075)\;
    $\mathit{results}[\mathit{config}]$ $\leftarrow$ $\mathit{CalculiX.Solve}$() $\rightarrow$ \{stress, displacement, weight\}\;
}

\tcp{PHASE 6: Intelligent Optimization (GP Regression + Bayesian Optimization)}
$\mathit{strategy}$ $\leftarrow$ $\mathit{OptimizationEngineer.AnalyzeTask}$($R$)\;
\If{$\mathit{strategy}$ == "$\mathit{large\_scale\_parameter\_sweep}$"}{
    $\mathit{GP\_stress}$ $\leftarrow$ $\mathit{GaussianProcess.Train}$($\mathit{struct\_matrix}$, $\mathit{results.stress}$)\;
    $\mathit{GP\_weight}$ $\leftarrow$ $\mathit{GaussianProcess.Train}$($\mathit{struct\_matrix}$, $\mathit{results.weight}$)\;

    $\mathcal{D}_{opt}$ $\leftarrow$ $\mathit{BayesianOptimization}$($\mathit{GP\_stress}$, $\mathit{GP\_weight}$, $\mathit{constraints}$)\;

    $\mathit{FEA\_optimal}$ $\leftarrow$ $\mathit{CalculiX.Solve}$($\mathcal{D}_{opt}$)\;
    $\mathit{improvement}$ $\leftarrow$ $\frac{\mathit{results.stress\_best} - \mathit{FEA\_optimal.stress}}{\mathit{results.stress\_best}}$\;
}

$\mathcal{M}$ $\leftarrow$ \{aerodynamic: \{L/D, OASPL\}, structural: \{stress, weight, $\mathit{improvement}$\}\}\;
\Return{$\mathcal{D}_{opt}$, $\mathcal{M}$}\;
\end{algorithm}

The optimized hybrid parallel scheduling strategy maximizes computational throughput while respecting resource constraints. By separating LLM-based planning (serial) from Docker-based CFD execution (parallel), the system achieves near-linear scaling up to the configured parallelism limit. The priority queue ensures that simulations with higher expected information gain are executed first, while the containerization provides isolation and reproducibility. This architecture enables the framework to evaluate hundreds of design configurations in hours rather than weeks, fundamentally transforming the engineering design cycle.

\section{Case Study: Multi-Agent Collaborative UAV Wing Optimization}

To demonstrate the framework's collaborative capabilities, we present a comprehensive case study showing how multiple engineering agents work together to optimize a UAV wing design.

\subsection{Problem Specification}

The natural language input to the system:

\begin{tcolorbox}[title=Prompt for UAV Case]
"Design a lightweight and efficient UAV wing for small drone applications. 
Focus on minimizing noise while maintaining good aerodynamic performance. 
Consider NACA series airfoils suitable for low Reynolds number operations 
($10^{5}$ to $10^{6}$). 
The wing should operate efficiently at cruise speeds of 
$25$--$35 \,\mathrm{m/s}$ with angles of attack from 
$0^{\circ}$ to $6^{\circ}$. 
Select the best wing model. 
Design a lightweight UAV wing with $100 \,\mathrm{mm}$ chord and 
$200 \,\mathrm{mm}$ span. 
Use aluminum 7075-T6. 
Minimize weight while ensuring safety factor $> 1.5$ under cruise,
maneuver, gust, and landing loads."
\end{tcolorbox}

\begin{figure}[H]
    \centering
    \includegraphics[width=1\linewidth]{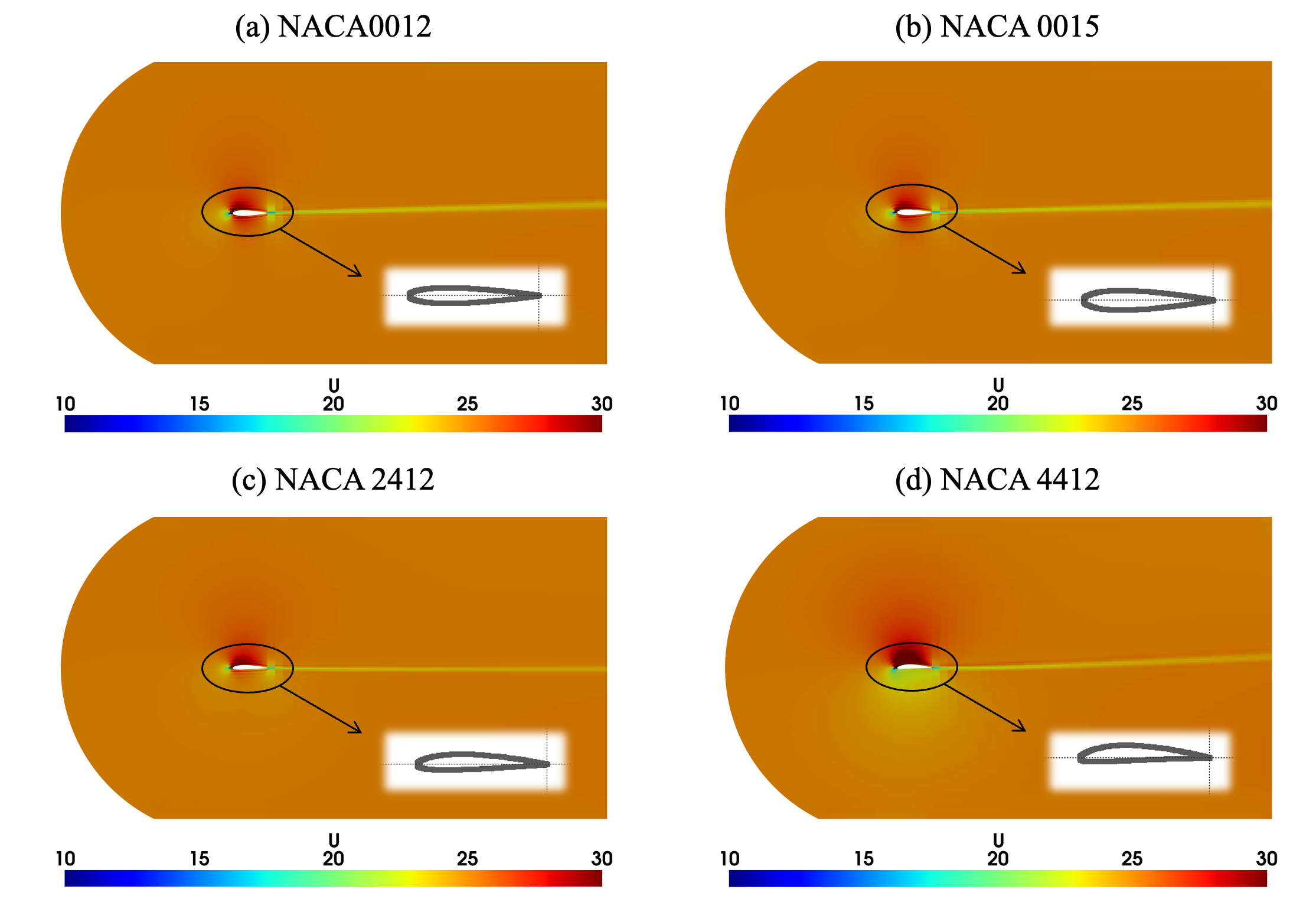}
    \caption{The four NACA airfoils selected by the Chief Engineer for UAV wing optimization. The design exploration encompasses symmetric profiles (NACA 0012, 0015) for baseline performance and cambered profiles (NACA 2412, 4412) for enhanced lift characteristics, systematically varying thickness (12\%-15\%) and camber (0\%-4\%) to explore the aerodynamic-structural-acoustic design space.}
    \label{fig:NACA}
\end{figure}

Upon analyzing the natural language requirements, the Chief Engineer initiated a comprehensive design exploration strategy. Recognizing the multi-objective nature of the UAV optimization problem—balancing aerodynamic efficiency, structural integrity, and acoustic performance—the system selected four NACA four-digit airfoils for detailed investigation, as illustrated in Figure~\ref{fig:NACA}.

The symmetric NACA 0012 airfoil serves as the baseline configuration due to its well-documented performance characteristics~\cite{Hassan2024E63Airfoil,abbott1959theory,jacobs1937naca}. NACA 0015 examines the effect of increased 15\% thickness. NACA 2412 and NACA 4412 investigate camber effects, both maintaining 12\% thickness but with 2\% and 4\% maximum camber respectively, located at 40\% chord position.

The geometric profiles of these NACA four-digit airfoils are generated using the analytical formulation, where the thickness distribution $y_t$ for a chord-normalized coordinate $x \in [0,1]$ is given by:

\begin{equation}
y_t = 5t \left[ 0.2969\sqrt{x} - 0.1260x - 0.3516x^2 + 0.2843x^3 - 0.1015x^4 \right]
\end{equation}

where $t$ represents the maximum thickness as a fraction of the chord. For cambered airfoils, the mean camber line $y_c$ is defined piecewise:

\begin{equation}
y_c = \begin{cases}
\frac{m}{p^2} \left( 2px - x^2 \right), & 0 \leq x \leq p \\
\frac{m}{(1-p)^2} \left[ (1-2p) + 2px - x^2 \right], & p < x \leq 1
\end{cases}
\end{equation}

where $m$ denotes the maximum camber and $p$ indicates the position of maximum camber along the chord. The final airfoil coordinates are obtained by distributing the thickness perpendicular to the camber line at each chordwise station.

For the specific airfoils in our benchmark suite:
\begin{itemize}
    \item \textbf{NACA 0012}: Symmetric profile with $m=0$, $p=0$, $t=0.12$
    \item \textbf{NACA 0015}: Symmetric profile with $m=0$, $p=0$, $t=0.15$
    \item \textbf{NACA 2412}: Cambered profile with $m=0.02$, $p=0.4$, $t=0.12$
    \item \textbf{NACA 4412}: Cambered profile with $m=0.04$, $p=0.4$, $t=0.12$
\end{itemize}

Through this systematic airfoil selection, the Chief Engineer established a comprehensive design matrix spanning Reynolds numbers from $10^{5}$ to $10^{6}$. The framework autonomously configured simulations across angles of attack from $0^{\circ}$ to $6^{\circ}$, ensuring coverage of both attached and separated flow conditions essential for understanding stall margins and control effectiveness. Operating velocities between $25$ and $35 \,\mathrm{m/s}$ were selected to remain within the incompressible flow assumptions while representing realistic UAV cruise and maneuvering speeds.

\subsection{Results and Performance}

Upon receiving the natural language specification, the \texttt{Engineering.ai} framework initiated a systematic multi-agent workflow, demonstrating the seamless integration of literature-based research, computational analysis, and autonomous decision-making.

\textbf{Phase 1: Literature Review and Experimental Design}

The Chief Engineer began by conducting comprehensive literature searches across multiple databases, retrieving 25 relevant publications on UAV aerodynamics and low-noise airfoil design. Through analysis of these literature, the system identified critical design parameters including Reynolds number, chord, and angle of attack. The Chief Engineer then orchestrated a multi-stage workflow: (1) aerodynamic analysis to select the optimal airfoil from candidate NACA profiles, (2) structural analysis by the Structural Engineer to explore key geometric design parameters, and (3) intelligent optimization by the Optimization Engineer to identify optimal designs balancing multiple performance objectives. For the initial aerodynamic stage, the Chief Engineer formulated a simulation matrix that would systematically explore the airfoil design space:

\begin{table}[H]
\centering
\caption{Simulation Matrix Designed by Chief Engineer}
\label{tab:simulation_matrix}
\begin{tabular}{cccccc}
\toprule
\textbf{Simulation} & \textbf{Airfoil} & \textbf{Chord (m)} & \textbf{Reynolds} & \textbf{Velocity (m/s)} & \textbf{AoA (°)} \\
\midrule
1 & NACA0012 & 0.10 & $2.91 \times 10^{5}$ & 25.0 & 0 \\
2 & NACA0012 & 0.10 & $3.50 \times 10^{5}$ & 30.0 & 3 \\
3 & NACA0012 & 0.10 & $4.08 \times 10^{5}$ & 35.0 & 6 \\
4 & NACA0015 & 0.10 & $2.91 \times 10^{5}$ & 25.0 & 2 \\
5 & NACA0015 & 0.10 & $3.50 \times 10^{5}$ & 30.0 & 4 \\
6 & NACA0015 & 0.10 & $4.08 \times 10^{5}$ & 35.0 & 1 \\
7 & NACA2412 & 0.10 & $2.91 \times 10^{5}$ & 25.0 & 1 \\
8 & NACA2412 & 0.10 & $3.50 \times 10^{5}$ & 30.0 & 5 \\
9 & NACA2412 & 0.10 & $4.08 \times 10^{5}$ & 35.0 & 2 \\
10 & NACA4412 & 0.10 & $2.91 \times 10^{5}$ & 25.0 & 5 \\
11 & NACA4412 & 0.10 & $3.50 \times 10^{5}$ & 30.0 & 0 \\
12 & NACA4412 & 0.10 & $4.08 \times 10^{5}$ & 35.0 & 4 \\
\bottomrule
\end{tabular}
\end{table}

This 12-simulation matrix represents the first stage of the multi-phase optimization pipeline. Upon completion of aerodynamic analysis and airfoil selection, the workflow would proceed to structural analysis (432 FEA configurations exploring geometric parameters) and intelligent optimization (GP-based surrogate modeling for design refinement).

\textbf{Phase 2: Aerodynamic Analysis}

The Aerodynamics Engineer autonomously executed all twelve simulation cases using OpenFOAM's \texttt{simpleFoam} solver with the Spalart-Allmaras turbulence model. The AI engineer employed Gmsh to generate structured C-type meshes with 40,000-45,000 nodes, followed by mesh quality verification using OpenFOAM's checkMesh utility and optimization using refineWallLayer. Boundary conditions were intelligently configured with velocity inlet components $\bigl(U \cos\alpha,\; U \sin\alpha,\; 0\bigr)$ to accurately represent angle-of-attack variations. The SIMPLE pressure-velocity coupling scheme~\cite{patankar1980numerical} utilized relaxation factors of 0.3 for pressure and 0.7 for velocity, automatically reduced to 0.2 and 0.5 respectively when detecting flow separation.

\begin{figure}[H]
    \centering
    \includegraphics[width=\linewidth]{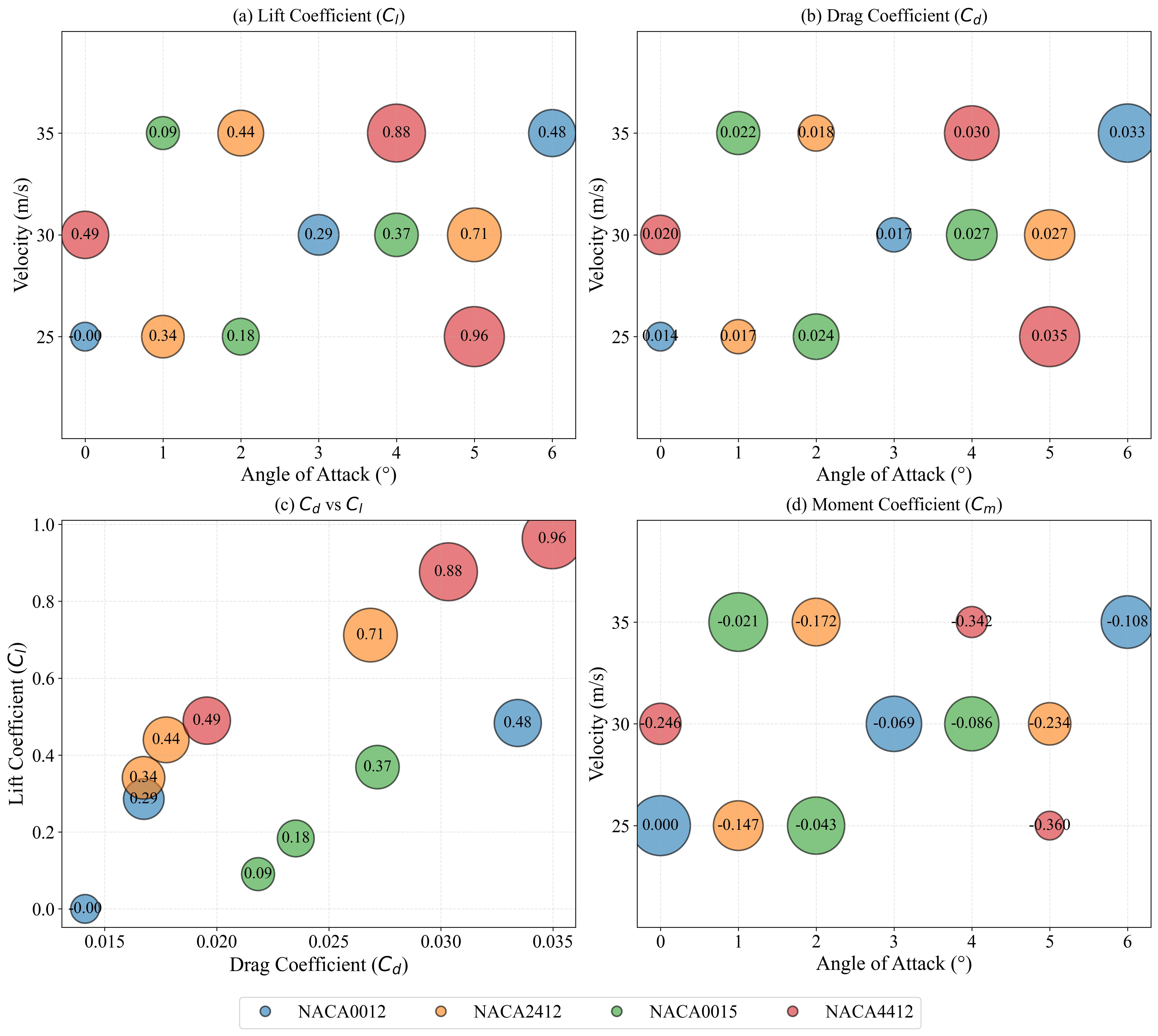}
    \caption{Comprehensive aerodynamic performance analysis across twelve simulation cases. (a)lift coefficient ($C_l$) variation with angle of attack and velocity, with bubble sizes proportional to $C_l$ magnitude. (b)drag coefficient ($C_d$) distribution across operating conditions with bubble sizes representing $C_d$ values. (c)drag polar ($C_d$ vs $C_l$) illustrating aerodynamic efficiency envelopes for each airfoil, with bubble sizes indicating $C_l$ magnitude. (d)pitching moment coefficient ($C_m$) characteristics with bubble sizes proportional to $|C_m|$.}
    \label{fig:aerodynamic_analysis}
\end{figure}

Figure~\ref{fig:aerodynamic_analysis} presents a comprehensive visualization of aerodynamic performance across twelve simulation cases at Reynolds numbers from $2.9 \times 10^{5}$ to $4.1 \times 10^{5}$. The upper-left panel reveals that NACA4412 achieved the highest lift coefficient of 0.96 at $5^{\circ}$ angle of attack and 25 m/s, demonstrating superior lift generation due to its 4\% camber. NACA0012 at $0^{\circ}$ exhibits near-zero lift as expected for a symmetric airfoil. The upper-right panel shows drag coefficients maintained below 0.035 across all conditions, with NACA0012 exhibiting the lowest drag of 0.014 at $0^{\circ}$ and 25 m/s. The drag polar in the lower-left panel illustrates distinct performance envelopes, with NACA4412 displaying a steep lift-drag relationship characteristic of highly cambered airfoils, while NACA0012 shows a flatter curve indicative of low-drag performance. Calculating lift-to-drag ratios reveals NACA4412 achieved maximum aerodynamic efficiency of 28.9 at $4^{\circ}$ and 35 m/s, followed by NACA2412 with 26.5 at $5^{\circ}$ and 30 m/s. The lower-right panel confirms characteristically negative pitching moments for cambered airfoils (NACA2412 and NACA4412), ranging from $-0.147$ to $-0.360$, providing inherent pitch stability for UAV applications, while symmetric airfoils (NACA0012 and NACA0015) exhibited near-zero moments.

\textbf{Phase 3: Acoustic Analysis}

Upon completion of the aerodynamic simulations, the Acoustic Engineer automatically extracted boundary-layer parameters from the CFD solutions, including displacement thickness $\delta^{*}$, momentum thickness $\theta$, and shape factor $H$ along the airfoil surface. Employing the Brooks-Pope-Marcolini (BPM) semi-empirical model, the system computed noise contributions from five distinct mechanisms: turbulent boundary-layer trailing-edge (TBL-TE) noise, separated-flow noise, laminar boundary-layer vortex-shedding noise, blunt trailing-edge vortex-shedding noise, and tip-vortex noise. Acoustic spectra were calculated from 100 Hz to 10 kHz at observer positions 1.0 m and 2.0 m perpendicular to the chord line.

\begin{figure}[H]
    \centering
    \includegraphics[width=\linewidth]{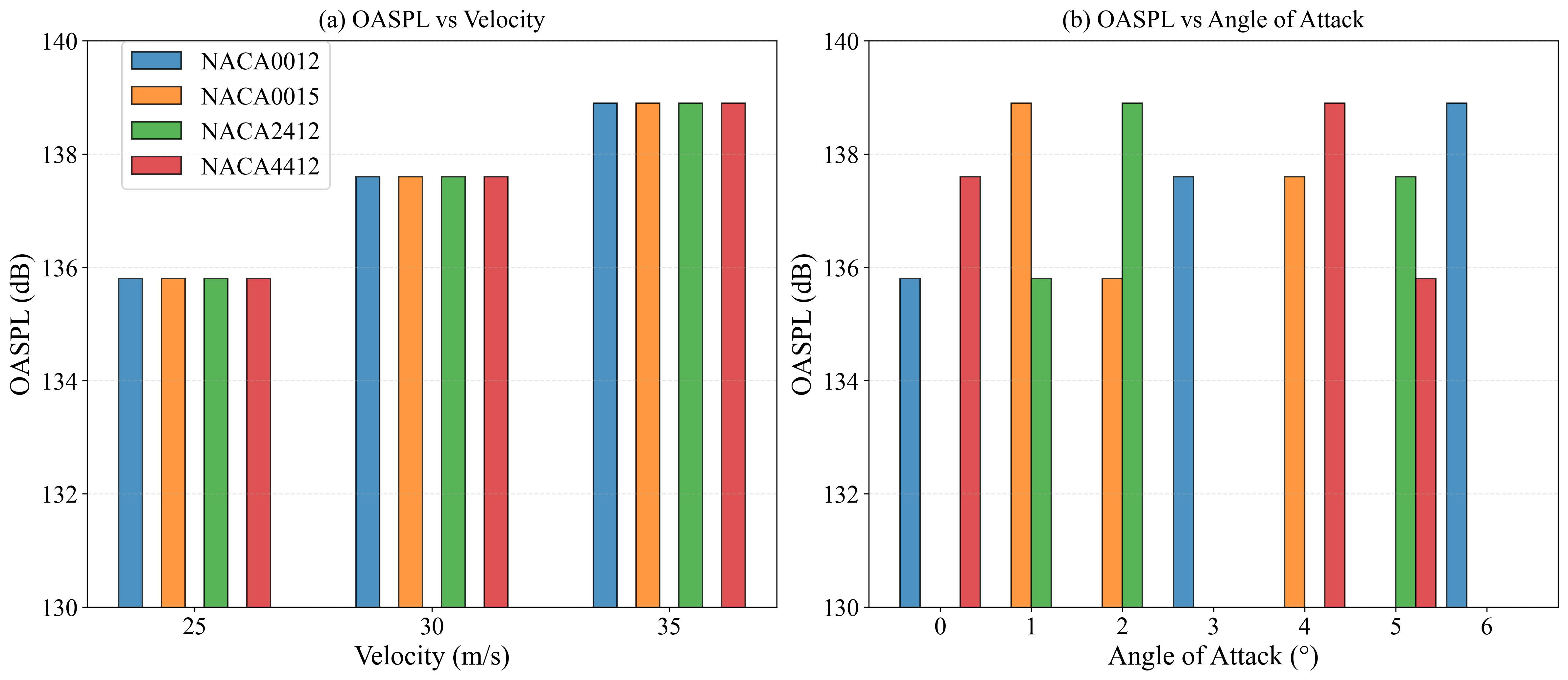}
    \caption{Acoustic performance comparison showing Overall Sound Pressure Level (OASPL) grouped by operating conditions. (a)OASPL variation with velocity (25-35 m/s), demonstrating identical acoustic levels across all airfoil geometries at each velocity. (b)OASPL distribution across angles of attack ($0^{\circ}$ to $6^{\circ}$), revealing that acoustic emissions are primarily velocity-dependent rather than geometry-dependent.}
    \label{fig:acoustic_analysis}
\end{figure}

The acoustic analysis revealed a remarkable finding: OASPL is primarily governed by velocity rather than airfoil geometry or angle of attack. At $25 \,\mathrm{m/s}$, all four airfoils (NACA0012, NACA0015, NACA2412, and NACA4412) exhibited identical OASPL values of $135.8 \,\mathrm{dB}$ regardless of their geometric differences or operating angles. This pattern persisted at higher velocities, with all configurations reaching $137.6 \,\mathrm{dB}$ at $30 \,\mathrm{m/s}$ and $138.9 \,\mathrm{dB}$ at $35 \,\mathrm{m/s}$. The consistent acoustic behavior across different airfoil geometries indicates that trailing-edge noise mechanisms dominate the acoustic signature, with boundary layer characteristics being primarily determined by Reynolds number rather than airfoil shape within this operating regime.

The observed velocity scaling of $3.1 \,\mathrm{dB}$ for a $40\%$ velocity increase (from 25 to 35 m/s) is consistent with dipole source scaling, where acoustic power scales approximately as $U^{5}$ to $U^{6}$, confirming that the noise generation is dominated by unsteady surface pressure fluctuations rather than quadrupole sources from turbulent mixing.

\textbf{Phase 4: Multi-Objective Performance Evaluation}

The Chief Engineer implemented a weighted performance metric combining aerodynamic efficiency and acoustic characteristics:

\begin{equation}
J = w_1 \cdot \frac{L/D}{(L/D)_{\max}} + w_2 \cdot \left(1 - \frac{\text{OASPL} - \text{OASPL}_{\min}}{\text{OASPL}_{\max} - \text{OASPL}_{\min}}\right)
\end{equation}

where $w_1 = 0.6$ prioritizes aerodynamic performance and $w_2 = 0.4$ accounts for noise reduction requirements. The comprehensive analysis revealed that acoustic performance is velocity-dominated with all airfoils exhibiting identical OASPL values at each velocity (135.8 dB at 25 m/s, 137.6 dB at 30 m/s, 138.9 dB at 35 m/s), effectively eliminating acoustic differentiation as a selection criterion. This finding shifted the optimization focus entirely to aerodynamic performance characteristics. Among the four configurations evaluated, NACA4412 emerged as the optimal selection based on multiple performance indicators. The airfoil achieved the maximum lift coefficient of 0.96 at $5^{\circ}$ and 25 m/s, and demonstrated the highest lift-to-drag ratio of 28.9 at $4^{\circ}$ and 35 m/s. Its 4\% camber provides superior lift generation across the operational envelope (0.49-0.96 $C_L$ range), crucial for stable UAV flight at varying speeds and attitudes. The boundary layer analysis revealed favorable shape factors ranging from 1.63 to 1.71, indicating excellent flow attachment and resistance to separation even at higher angles of attack. The cambered profile's inherent pitch stability, evidenced by consistent negative pitching moments ($C_m$ = $-0.246$ to $-0.360$), eliminates the need for aggressive control surface deflections, reducing induced drag and improving overall efficiency. This selection prioritizes predictable lift generation, superior aerodynamic efficiency, and operational stability over the marginally lower drag of symmetric airfoils at zero lift conditions.

\textbf{Phase 5: Structural Analysis}

\begin{figure}[!htb]
    \centering
    \includegraphics[width=\textwidth]{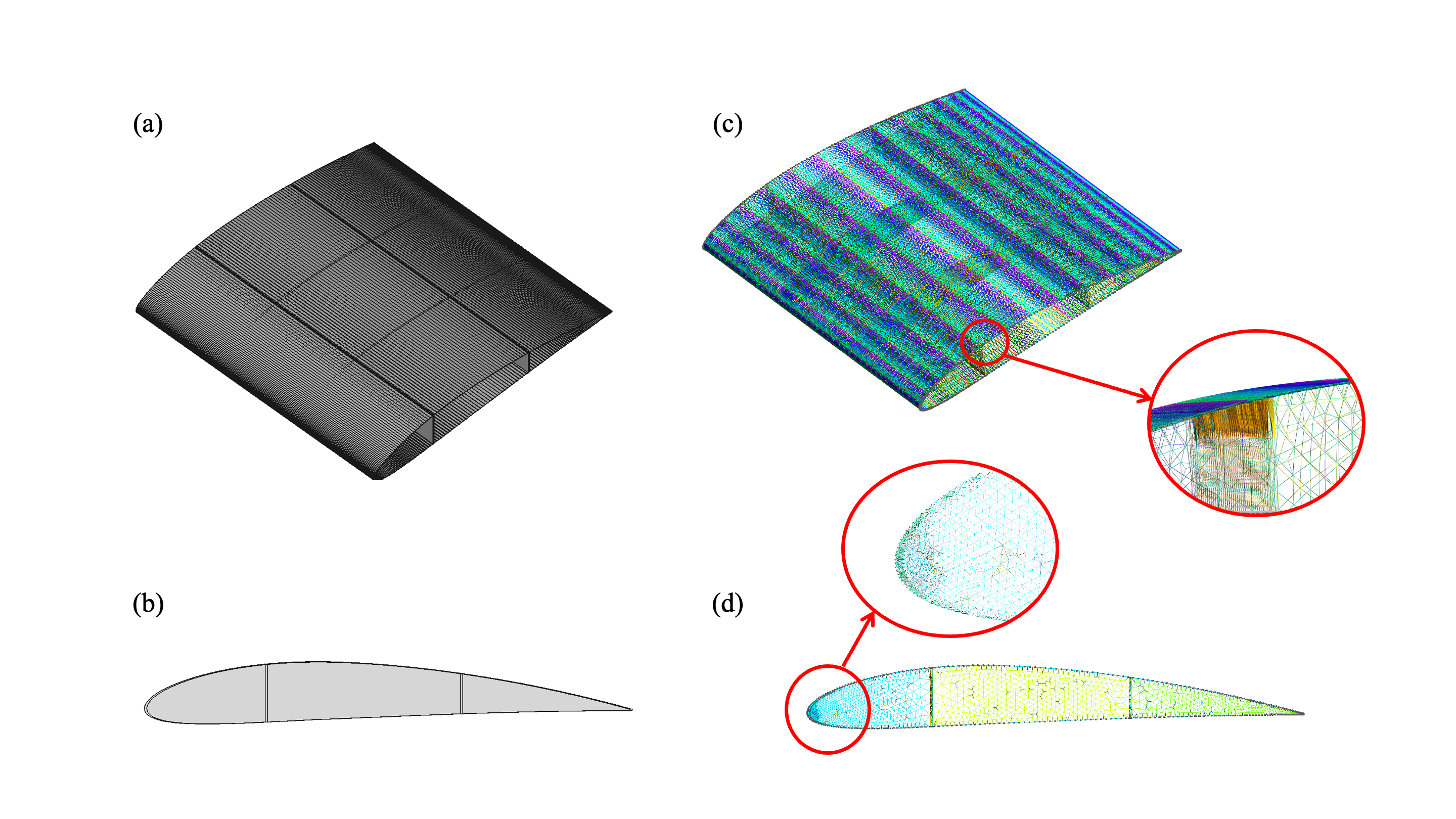}
    \caption{Autonomous CAD-to-FEA workflow by the Structural Engineer: \textbf{(a)} Parametric wing geometry in FreeCAD with internal structure. \textbf{(b)} NACA 4412 cross-section. \textbf{(c)} Adaptive tetrahedral mesh by Gmsh with local refinement. \textbf{(d)} Cross-sectional mesh view. This automated pipeline enabled exploration of 432 structural configurations.}
    \label{fig:cad_mesh}
\end{figure}

Following the NACA 4412 airfoil selection, the Structural Engineer autonomously initiated a comprehensive structural design exploration across 432 configurations through an integrated CAD-to-FEA workflow. The automated pipeline (Figure~\ref{fig:cad_mesh}) begins with FreeCAD-based parametric geometry generation, where the agent creates 3D wing models incorporating the selected NACA 4412 airfoil profile with internal structural components including spars (longitudinal stiffeners) and ribs (spanwise stiffeners). The parametric approach enables systematic variation of key structural parameters: spar width (0.2--2.0 mm), rib thickness (0.5--2.0 mm), shell thickness (1.0--3.0 mm), number of spars (2--3), and number of ribs (2--3). The Structural Engineer autonomously determined these parameter ranges based on engineering constraints and manufacturability considerations. The lower bounds (0.2 mm spar width, 0.5 mm rib thickness, 1.0 mm shell thickness) represent minimum manufacturable dimensions for typical UAV fabrication processes, while upper bounds (2.0 mm, 2.0 mm, 3.0 mm) are constrained by weight budget and structural proportions. The discrete parameter sweep generated 432 unique configurations ($3 \times 6 \times 6 \times 2 \times 2 = 432$), providing sufficient design space coverage to capture nonlinear stress-geometry relationships while maintaining computational tractability.

The FreeCAD-generated STEP geometry files are subsequently processed by Gmsh for finite element mesh generation. The agent employs adaptive tetrahedral meshing~\cite{shewchuk1996triangle,field1988laplacian} with second-order elements (10-node tetrahedra) to accurately capture stress gradients. Local mesh refinement is automatically applied near geometric discontinuities such as spar-shell junctions, rib-shell interfaces, and leading/trailing edges, where stress concentrations are anticipated. This adaptive strategy balances computational efficiency with solution accuracy across the diverse geometric configurations.

\begin{figure}[!htb]
    \centering
    \includegraphics[width=\textwidth]{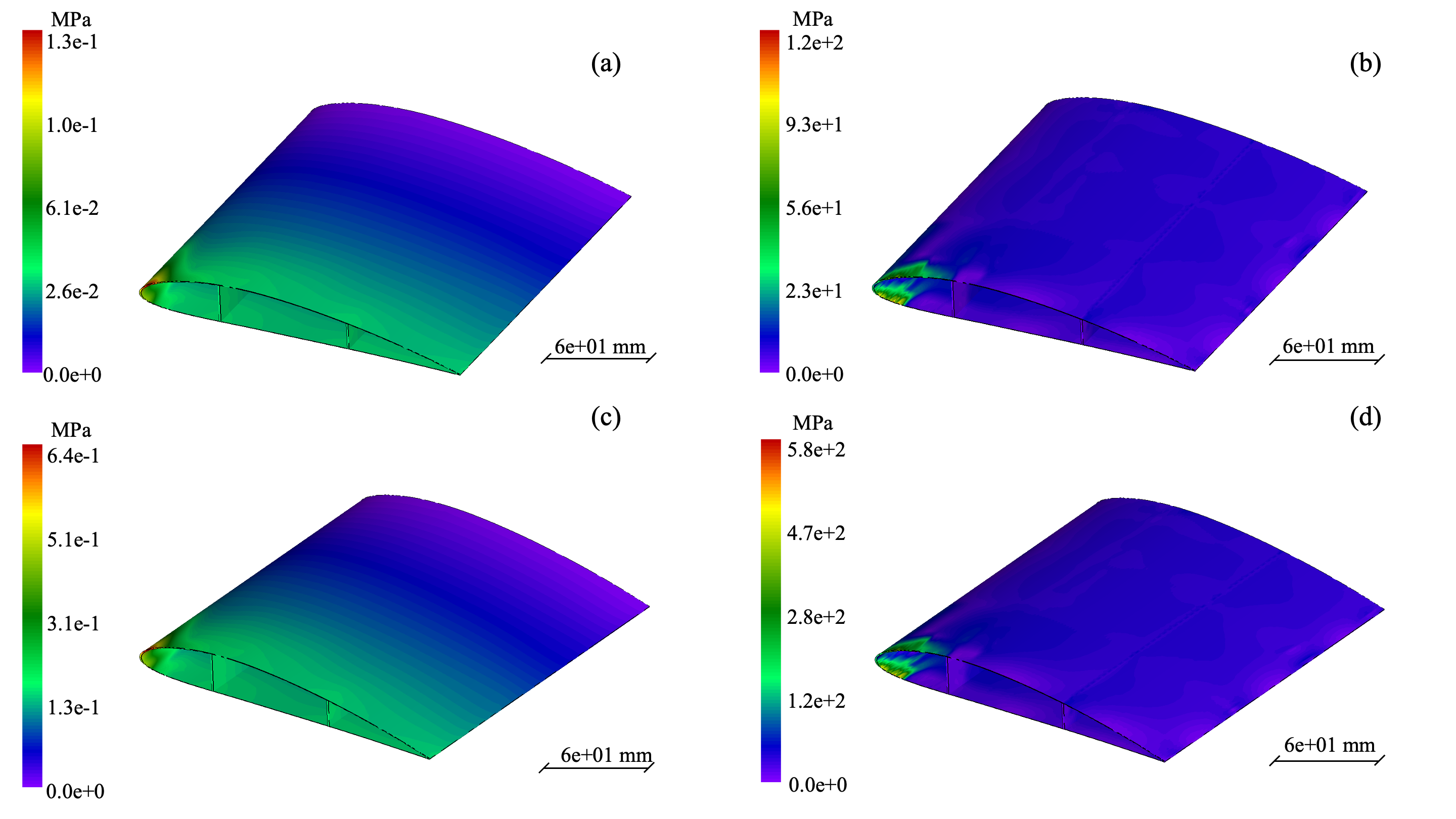}
    \caption{Representative CalculiX finite element analysis results from the Structural Engineer's 432-configuration parameter sweep. \textbf{(a)} Displacement field under cruise load (STEP 1, 1g). \textbf{(b)} von Mises stress distribution under cruise load. \textbf{(c)} Displacement field under landing impact load (STEP 4, 3g vertical). \textbf{(d)} von Mises stress distribution under landing impact load. The stress concentration patterns (color gradients from green to purple) at spar-shell junctions reveal critical load paths that guided subsequent optimization.}
    \label{fig:calculix_fea}
\end{figure}

For finite element analysis, CalculiX is configured with aluminum 7075-T6 material properties~\cite{asm1990properties} (E=71.7 GPa, $\nu$=0.33, $\rho$=2810 kg/m³). The agent autonomously applies realistic flight load conditions including cruise aerodynamic loads (inherited from Phase 1-4), maneuver load factors, gust loads, and landing impact scenarios. Boundary conditions enforce root fixity to represent wing-fuselage attachment. The von Mises stress distributions from representative configurations (Figure~\ref{fig:calculix_fea}) reveal critical load paths concentrated at spar-shell junctions and rib attachment points, providing physical insights that guide subsequent optimization. Remarkably, the automated CAD-to-FEA pipeline achieved a 100\% success rate across all 432 configurations, with zero mesh generation failures, convergence issues, or manual interventions required. This robustness validates the Structural Engineer's autonomous error-handling capabilities and adaptive meshing strategies. Upon completion of all 432 FEA simulations, the Structural Engineer transmitted the complete dataset—including stress fields, displacement magnitudes, and structural mass—to the Optimization Engineer for subsequent optimization and data analysis.

\textbf{Phase 6: Intelligent Optimization}

Upon receiving the complete structural dataset from the Structural Engineer, the Optimization Engineer—specialized in machine learning-based design space exploration—initiated multi-objective optimization to discover superior designs beyond the discrete parameter sweep. The agent autonomously selected Gaussian Process (GP) regression~\cite{rasmussen2006gaussian} as the surrogate modeling strategy to address the computational cost of exhaustive parameter search. For each response variable $y$ (stress or weight), the GP model assumes:
\begin{equation}
y(\mathbf{x}) \sim \mathcal{GP}(\mu(\mathbf{x}), k(\mathbf{x}, \mathbf{x}'))
\end{equation}
where $\mu(\mathbf{x})$ is the mean function and $k(\mathbf{x}, \mathbf{x}')$ is the RBF covariance function $k(\mathbf{x}, \mathbf{x}') = \sigma_f^2 \exp(-\frac{||\mathbf{x}-\mathbf{x}'||^2}{2\ell^2})$ with length-scale $\ell$ and signal variance $\sigma_f^2$. Two independent GP models were trained for stress and weight prediction using 80\% of the FEA data (345 samples) for training and 20\% (87 samples) for validation.

\begin{figure}[!htb]
    \centering
    \includegraphics[width=\textwidth]{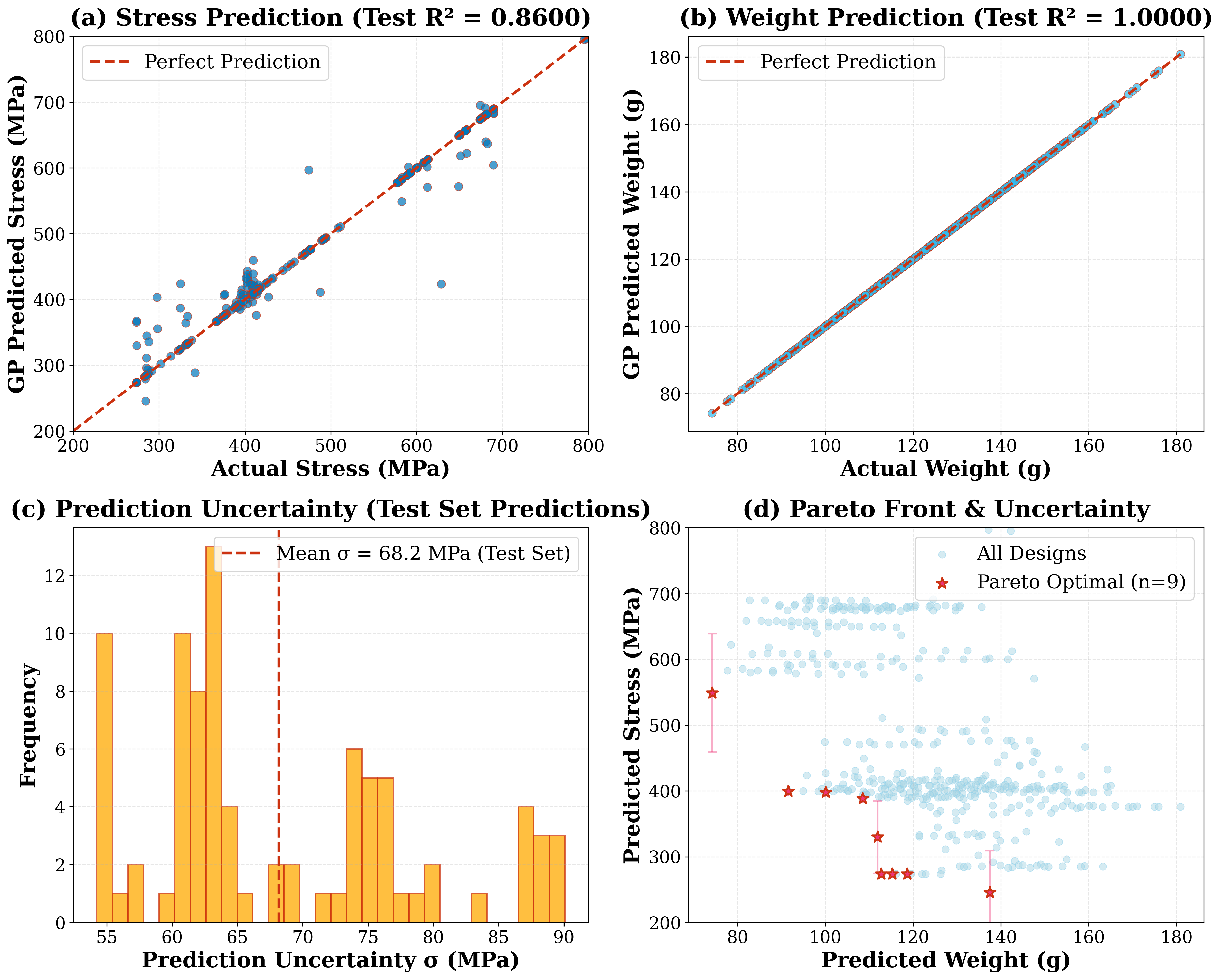}
    \caption{Gaussian Process-based optimization by the Optimization Engineer. \textbf{(a)} Stress prediction accuracy ($R^2=0.86$). \textbf{(b)} Weight prediction accuracy ($R^2=1.00$). \textbf{(c)} Prediction uncertainty distribution. \textbf{(d)} Pareto front analysis with nine optimal designs. The continuous optimization discovered a design with 18.1\% stress reduction compared to the best discrete configuration.}
    \label{fig:gp_optimization}
\end{figure}

The GP models achieved strong predictive performance (Figure~\ref{fig:gp_optimization}a,b) with $R^2=0.86$ for stress and $R^2=1.00$ for weight on test data. The prediction uncertainty analysis (Figure~\ref{fig:gp_optimization}c) reveals spatial variation in model confidence, with mean uncertainty $\sigma=68.2$ MPa. Low-uncertainty regions ($\sigma<70$ MPa, 55.2\% of predictions) correspond to well-explored parameter combinations from the FEA dataset, while high-uncertainty regions ($\sigma>80$ MPa, 13.8\%) identify areas where the surrogate model extrapolates beyond training data. This uncertainty quantification enables principled exploration-exploitation trade-offs through the Expected Improvement (EI) acquisition function~\cite{jones1998efficient,mockus1974bayesian}:
\begin{equation}
\text{EI}(\mathbf{x}) = \mathbb{E}[\max(f_{\text{best}} - f(\mathbf{x}), 0)]
\end{equation}
where $f_{\text{best}}$ is the current best objective value and $f(\mathbf{x})$ is the GP-predicted response at candidate design $\mathbf{x}$.

The multi-objective optimization problem~\cite{deb2002fast} is formulated as:
\begin{equation}
\begin{aligned}
\text{minimize} \quad & f_1(\mathbf{x}) = \sigma_{\text{stress}}(\mathbf{x}) \\
\text{minimize} \quad & f_2(\mathbf{x}) = m(\mathbf{x}) \\
\text{subject to} \quad & \mathbf{x}_{\text{min}} \leq \mathbf{x} \leq \mathbf{x}_{\text{max}}
\end{aligned}
\end{equation}
where $\sigma_{\text{stress}}$ is the von Mises stress and $m$ is the structural mass. The Pareto front analysis (Figure~\ref{fig:gp_optimization}d) identified nine optimal designs spanning the stress-weight trade-off space. Critically, the continuous Bayesian optimization discovered a superior design with \textbf{SPAR\_WIDTH}=1.902 mm, \textbf{RIB\_THICKNESS}=1.717 mm, and \textbf{SHELL\_THICKNESS}=1.592 mm—parameter values between discrete grid points that would not be evaluated by traditional methods. This configuration achieves predicted stress of $224\pm51$ MPa at $125.9\pm0.01$ g, representing an 18.1\% stress reduction compared to the best discrete design (274 MPa at 125.4 g).

\subsection{Performance on efficiency and cost}

The comprehensive evaluation of \texttt{engineering.ai} demonstrates transformative performance improvements in computational engineering workflows. The entire process from initial design through simulation to final optimization can now be completed in 2-3 hours. This acceleration fundamentally changes the feasibility of comprehensive design exploration within practical project constraints.

The framework achieves these improvements through optimized resource utilization and intelligent task scheduling. By separating LLM-based planning (serial operations) from Docker-containerized CFD execution (parallel operations), CPU utilization increases from 25-30\% in traditional workflows to 70-85\%. The system allocates 8-10 GB memory for parallel execution compared to 2-3 GB for single-task processing, achieving 5$\times$ throughput improvement with only 3.3× memory overhead. This architecture enables simultaneous execution of multiple OpenFOAM simulations while maintaining compliance with API rate limitations.

Error recovery constitutes a critical capability of the autonomous framework. The system implements context-aware diagnosis and correction for common failure modes. Mesh quality issues, the most frequent failure type, trigger adaptive refinement adjustments based on geometric complexity. Solver convergence problems automatically modify relaxation factors (pressure: 0.7→0.3, velocity: 0.5→0.2) and time-stepping schemes. Boundary condition errors are resolved through pattern-based detection and domain-specific corrections. The checkpointing system enables recovery with less than 2\% overhead while maintaining state consistency.

\section{Discussion}

\texttt{Engineering.ai} represents a fundamental paradigm shift from tool-specific scripting to intelligent, autonomous engineering computation. The LLM's ability to understand physics, make engineering judgments, and learn from failures transforms how engineers interact with simulation tools.

The framework's natural language interface allows engineers to describe problems in plain language and receive optimized designs without writing a single line of code. Through sophisticated LLM reasoning, most simulation errors can be automatically resolved without the need for human intervention, from mesh quality issues to solver convergence problems. \texttt{Engineering.ai} provides an explicit control flow that ensures reproducible results and debuggable execution paths. Although the explicit control flow design sacrifices the creativity and adaptability of fully autonomous agents in unforeseen scenarios, it ensures the reproducibility and reliability essential for production engineering workflows.


Despite these capabilities, several challenges remain. API rate limits highlight the mismatch between parallel computational workflows and sequential API processing. Despite the implementation of various scheduling strategies, the inability to achieve true parallel LLM execution indicates that current API architectures are not optimized for computational engineering workflows. Each simulation case requires 15,000–20,000 tokens, which may become prohibitively expensive for large-scale industrial deployments involving thousands of design iterations. Privacy and intellectual property concerns also arise when transmitting proprietary design specifications to external APIs, underscoring the need for local LLM deployment options. The observed performance degradation when using smaller open-source models reveals a strong dependence on cutting-edge model capabilities, raising questions about long-term sustainability as model performance and pricing evolve.

\texttt{Engineering.ai}’s success in automating routine tasks suggests a future in which engineers focus on problem formulation, constraint definition, and result interpretation rather than software operation. However, the aspects that still require human intervention often involve the most interesting and valuable engineering insights, underscoring that human intuition remains irreplaceable for innovative design. Democratizing CFD tools through natural-language interfaces can expand the user base beyond domain experts, but it also raises concerns about result interpretation without a deep understanding of the underlying physics and numerical methods. LLM hallucination occasionally produces physically implausible suggestions, necessitating robust validation mechanisms to ensure solution reliability. The computational overhead of LLM API calls can make the framework less efficient than traditional methods for simple, well-defined problems. The black-box nature of some LLM decisions poses challenges for critical applications where full traceability and explainability are required. Currently limited to open-source tools like FreeCAD, Gmsh, and CalculiX, the framework requires extension to commercial software packages to achieve broader industry adoption. These limitations, however, are not fundamental barriers but rather engineering challenges that will be addressed as the technology matures.

\section{Conclusions}

\texttt{Engineering.ai} establishes autonomous computational engineering through a hierarchical multi-agent architecture where specialized AI engineers collaborate under Chief Engineer coordination. The system transforms natural language requirements into executable workflows, autonomously managing geometry generation, mesh optimization, multidisciplinary analysis, and design optimization.

The UAV wing optimization case demonstrates comprehensive autonomous capabilities. Starting from specifications for lightweight, efficient wings operating at $25$ to $35 \,\mathrm{m/s}$, the Chief Engineer formulated a simulation matrix spanning four NACA airfoils at Reynolds numbers from $2.91\times10^{5}$ to $4.08\times10^{5}$. The Aerodynamics Engineer executed parallel CFD analyses using \texttt{simpleFoam} with Spalart--Allmaras turbulence modeling. The Acoustic Engineer applied BPM analysis, revealing velocity-dominated noise ($135.8 \,\mathrm{dB}$ at $25 \,\mathrm{m/s}$, $137.6 \,\mathrm{dB}$ at $30 \,\mathrm{m/s}$, $138.9 \,\mathrm{dB}$ at $35 \,\mathrm{m/s}$), independent of airfoil geometry. Comprehensive analysis identified NACA 4412 as optimal, achieving a maximum lift coefficient of $0.96$ at $5^{\circ}$ and a superior lift-to-drag ratio of $28.9$ at $4^{\circ}$ and $35 \,\mathrm{m/s}$. The Structural Engineer autonomously executed $432$ configurations through an integrated FreeCAD-Gmsh-CalculiX pipeline under cruise, maneuver, gust, and landing loads, achieving $100\%$ success rate with zero failures or manual interventions. Stress analysis revealed design space spanning $224$--$680 \,\mathrm{MPa}$ and $78$--$178 \,\mathrm{g}$ across parametric variations in spar width, rib thickness, and shell thickness. The Optimization Engineer employed Gaussian Process surrogate modeling ($R^{2}=0.86$ for stress, $R^{2}=1.00$ for weight) with Bayesian optimization to discover an optimal design achieving $18.1\%$ stress reduction ($224 \,\mathrm{MPa}$ at $125.9 \,\mathrm{g}$) compared to the best discrete configuration, with nine Pareto-optimal solutions identified spanning the stress-weight trade-off space.

For this specific case, setup and iteration times of \texttt{Engineering.ai} have been significantly reduced, transforming processes that traditionally required days into minutes of execution. The architectural choices embodied in \texttt{Engineering.ai} provide valuable lessons for future AI-driven engineering systems. They prioritize reliability and debuggability, which are critical in engineering applications, with reproducibility as the foremost requirement. The separation of LLM-based reasoning from numerical computation, with well-defined interfaces and validation checkpoints, enables robust error handling and incremental enhancement.

\texttt{Engineering.ai} holds the potential to transform how we approach computational engineering design, democratize access to advanced engineering capabilities, and accelerate innovation cycles to an unprecedented degree. Our work indicates a path toward a new collaborative era where AI engineers work alongside human engineers, each complementing the other’s strengths to solve humanity’s most challenging engineering problems.

\section*{Author Contributions}

\textbf{Ran Xu}: System architecture design, multi-agent framework development, OpenFOAM integration for fluid dynamics simulations, acoustic simulation and analysis implementation, UAV wing optimization case design and implementation, manuscript writing and revision.

\textbf{Yupeng Qi}: System architecture design, technical implementation of the CAD/CAE pipeline, FreeCAD/Gmsh/CalculiX integration, machine learning optimization algorithms, experimental validation, manuscript writing and revision. Developed the optimization design modules and parameter exploration framework.

\textbf{Jingsen Feng}: Manuscript revision, data verification and validation, technical review of simulation results, and quality assurance throughout the research process.

\textbf{Xu Chu}: Conceptualization and original idea, project supervision, funding acquisition, strategic direction, code architecture optimization, algorithm design, manuscript writing and revision. Provided the foundational vision for autonomous AI engineering teams and guided the framework's development toward cross-domain applications.

All authors contributed to the manuscript preparation and participated in critical discussions that shaped the final framework design.

\section*{Competing Interests}

The authors declare no competing financial or non-financial interests.

\section*{Acknowledgments}

This work was supported by China Scholarship Council (CSC).

\appendix

\section{Validation of Autonomous Engineering Agents}

To establish the credibility and accuracy of the autonomous engineering framework, rigorous validation studies were conducted for both the Aerodynamics Engineer and Acoustic Engineer agents against established experimental datasets and benchmark simulations.

\subsection{Aerodynamics Engineer Validation}

The Aerodynamics Engineer's predictions were validated against experimental data from Ladson et al.~\cite{ladson1988effects} for the NACA 0012 airfoil at Reynolds number $Re = 6 \times 10^{6}$. The validation encompassed three critical aerodynamic metrics: drag coefficient ($C_d$), lift coefficient ($C_l$), and pressure coefficient distribution ($C_p$) at multiple angles of attack.

\begin{figure}[H]
    \centering
    \includegraphics[width=1\textwidth]{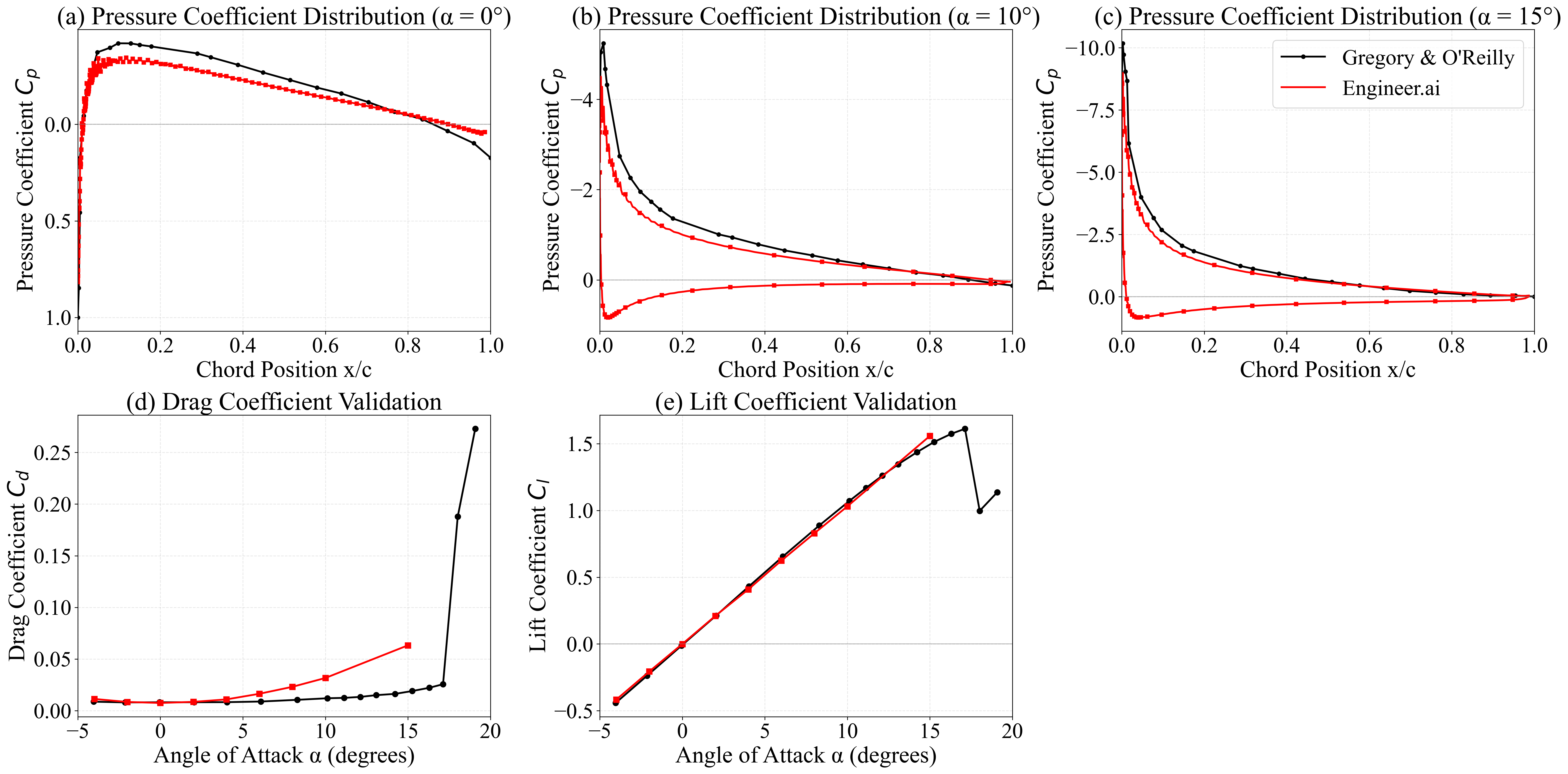}
    \caption{Aerodynamic validation showing drag coefficient, lift coefficient, and pressure distributions at three angles of attack ($0^{\circ}$, $10^{\circ}$, $15^{\circ}$). Engineer.ai predictions (red) compared with Ladson experimental data (black). Validation against Ladson et al. wind tunnel measurements for NACA 0012 at $Re = 6 \times 10^{6}$.}
    \label{fig:aerodynamic_validation}
\end{figure}

Figure~\ref{fig:aerodynamic_validation} demonstrates the comparison between framework predictions and experimental measurements. For drag coefficient validation, the framework captured the low-drag characteristic of symmetric airfoils at small angles of attack ($-4^{\circ}$ to $4^{\circ}$), with absolute errors below 0.003. The lift coefficient predictions showed good linear correlation with experimental data from $-4^{\circ}$ to $10^{\circ}$ angle of attack. At higher angles approaching stall ($15^{\circ}$), predictions showed larger deviations due to the challenges of modeling transitional and separated flow using RANS turbulence models.

Pressure coefficient distributions were validated at three representative angles of attack: $0^{\circ}$, $10^{\circ}$, and $15^{\circ}$. At $0^{\circ}$, the framework captured the symmetric pressure distribution trend. At $10^{\circ}$, the pressure distribution patterns were reasonably reproduced. At $15^{\circ}$ approaching stall, larger discrepancies emerged as expected when modeling complex separated flow conditions with RANS-based approaches.

\subsection{Acoustic Engineer Validation}

The Acoustic Engineer agent was validated against the Airfoil Self-Noise dataset from the UCI Machine Learning Repository~\cite{Dua:2019}, which provides benchmark acoustic measurements for NACA 0012 airfoil obtained from NASA wind tunnel experiments under controlled conditions. Five representative test cases were selected spanning velocities from 31.7 m/s to 55.5 m/s and angles of attack from $0.0^{\circ}$ to $6.7^{\circ}$.

\begin{figure}[H]
    \centering
    \includegraphics[width=1\textwidth]{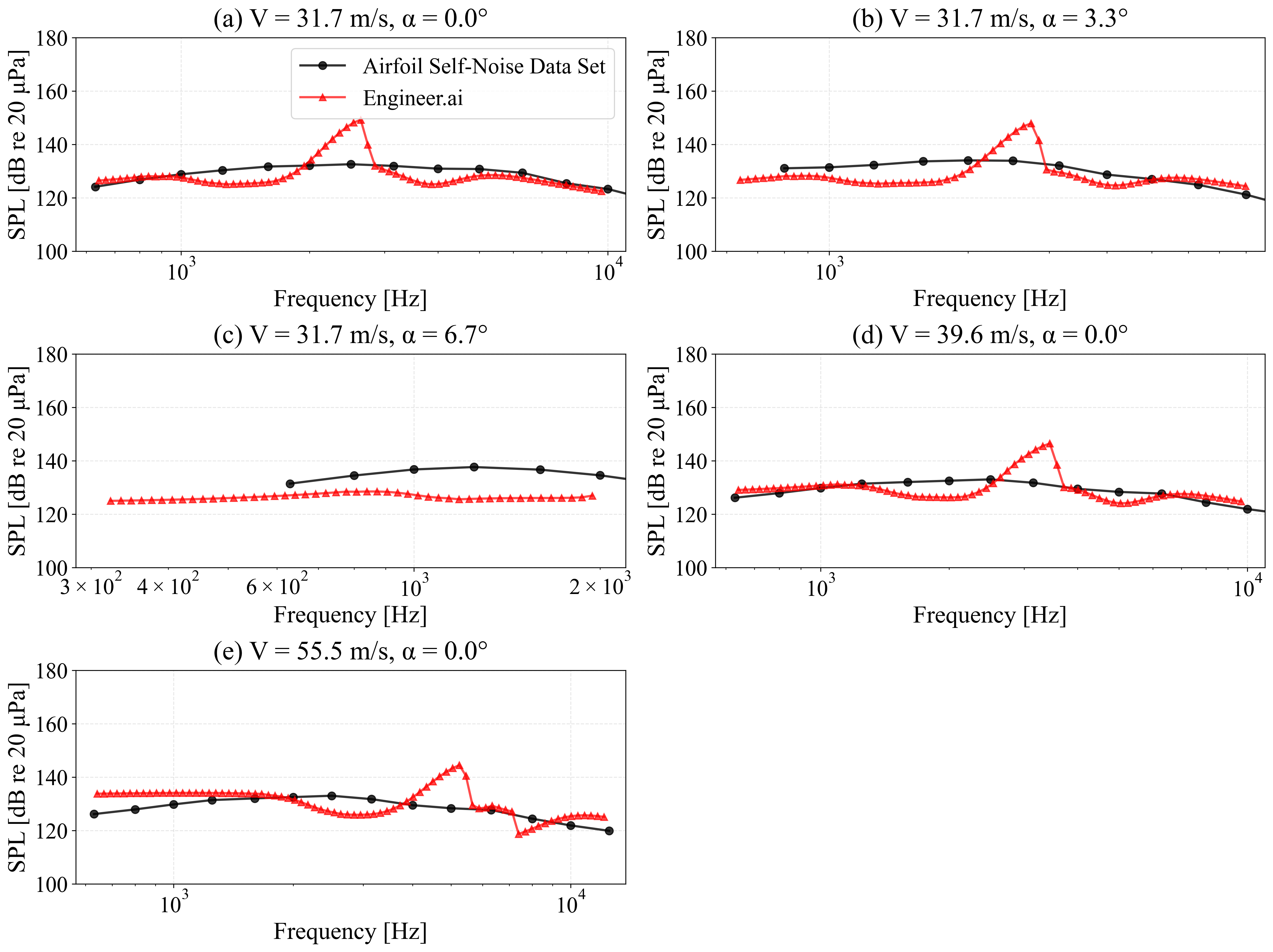}
    \caption{Acoustic validation comparing predicted SPL spectra against NASA Airfoil Self-Noise Database measurements at five operating conditions. The BPM-based predictions show RMSE ranging from 4.2 to 6.3 dB across all cases. Validation against Airfoil Self-Noise dataset from UCI Machine Learning Repository~\cite{Dua:2019} spanning velocities 31.7--55.5 m/s and angles of attack $0.0^{\circ}$--$6.7^{\circ}$.}
    \label{fig:acoustic_validation}
\end{figure}

Figure~\ref{fig:acoustic_validation} presents the comparison of predicted sound pressure level (SPL) spectra against experimental measurements across the frequency range of 315 Hz to 12.5 kHz. The Brooks-Pope-Marcolini (BPM) model implementation within the Acoustic Engineer demonstrates reasonable predictive capability, with root-mean-square errors (RMSE) ranging from 4.2 to 6.3 dB across all test conditions. The framework captured the general spectral trends, including the broadband nature of turbulent boundary layer trailing-edge noise.

At the baseline condition ($V = 31.7$ m/s, $\alpha = 0.0^{\circ}$), the predicted SPL spectrum showed agreement with experimental data with RMSE of 6.0 dB across the 630 Hz to 10 kHz frequency range. The framework successfully captured the increase in acoustic levels with velocity. At higher angles of attack ($\alpha = 6.7^{\circ}$), the predictions showed RMSE of 4.2 dB over the 315 Hz to 2 kHz range, confirming the autonomous acoustic analysis provides useful engineering predictions within acceptable accuracy for preliminary design applications.

These validation studies demonstrate that the autonomous engineering agents provide useful predictions for preliminary engineering design applications. The results show reasonable agreement in attached flow regimes, with expected limitations in complex separated flow conditions typical of RANS-based CFD approaches.

\bibliography{bibliography}

\end{document}